\documentclass[10pt]{amap-modern}

\usepackage{amsmath}
\usepackage{amssymb}
\usepackage{tabularx}
\usepackage{enumitem}
\usepackage{xspace}
\usepackage{wrapfig}

\setlist[itemize]{leftmargin=1.5em}

\makeatletter
\DeclareRobustCommand\onedot{\futurelet\@let@token\@onedot}
\def\@onedot{\ifx\@let@token.\else.\null\fi\xspace}

\makeatother

\newcommand{\method}{\textit{ABot-Recon}\xspace}

\crefname{figure}{Fig.}{Figs.}
\Crefname{figure}{Fig.}{Figs.}
\crefname{table}{Tab.}{Tabs.}
\Crefname{table}{Tab.}{Tabs.}
\crefname{section}{Sec.}{Secs.}
\Crefname{section}{Sec.}{Secs.}
\crefname{equation}{Eq.}{Eqs.}
\Crefname{equation}{Eq.}{Eqs.}

\headerlogo{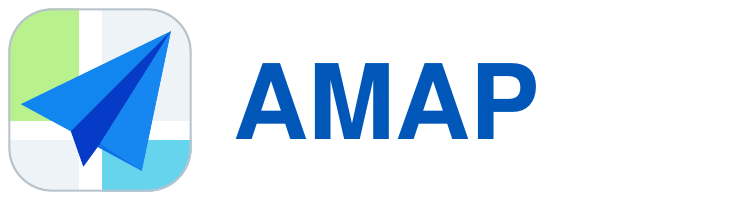}

\title{Revisiting Local Context for Long-Horizon Streaming 3D Reconstruction}

\author{AMAP CV Lab}
\affiliation{Alibaba Group}

\abstract{%

Streaming 3D reconstruction from extremely long videos requires estimating
camera motion and scene geometry online under bounded memory and computation.
Early streaming models achieve causal, bounded-cost inference using finite
context buffers or compact recurrent states, yet their estimates often
deteriorate as sequences grow. Recent methods improve long-horizon stability by
coupling short-range context with persistent or multi-level long-range memory.
We pursue a different route: we keep the learned temporal state strictly local
and formulate predictions whose targets remain independent of sequence length.
We present \method{}, a simple streaming model that caches KV features from only
the preceding 11 frames. It predicts a point map in the current camera
coordinate system together with an adjacent-frame relative pose. These
predictions remain equivariant under changes of reference frame, and global
poses and geometry are recovered through sequential composition. To reduce
accumulated drift, a lightweight temporal refiner improves relative rotations using recent visual and motion context, while a composition-aware pose loss supervises multi-step pose composition.
Extensive evaluations on challenging long-sequence benchmarks demonstrate the superior long-horizon performance of our local-context approach. On Oxford Spires, \method{} achieves an ATE of 4.35 m and an RPE-R of $0.12^\circ$, reducing both errors by approximately 40\% relative to the best prior results.

\par\smallskip
\noindent
\textbf{Project Page:} \url{https://amap-cvlab.github.io/ABot-Recon-html}
\noindent

\textbf{Code:} \url{https://github.com/amap-cvlab/ABot-Recon}
}

\begin{document}

\maketitle

\section{Introduction}
\label{sec:introduction}

\setlength{\intextsep}{0pt}   
\setlength{\columnsep}{10pt}  

\begin{wrapfigure}[18]{r}{0.55\textwidth}
  \vspace{-0.8\baselineskip}
  \centering

  \includegraphics[
    width=\linewidth,
    height=6.0cm
  ]{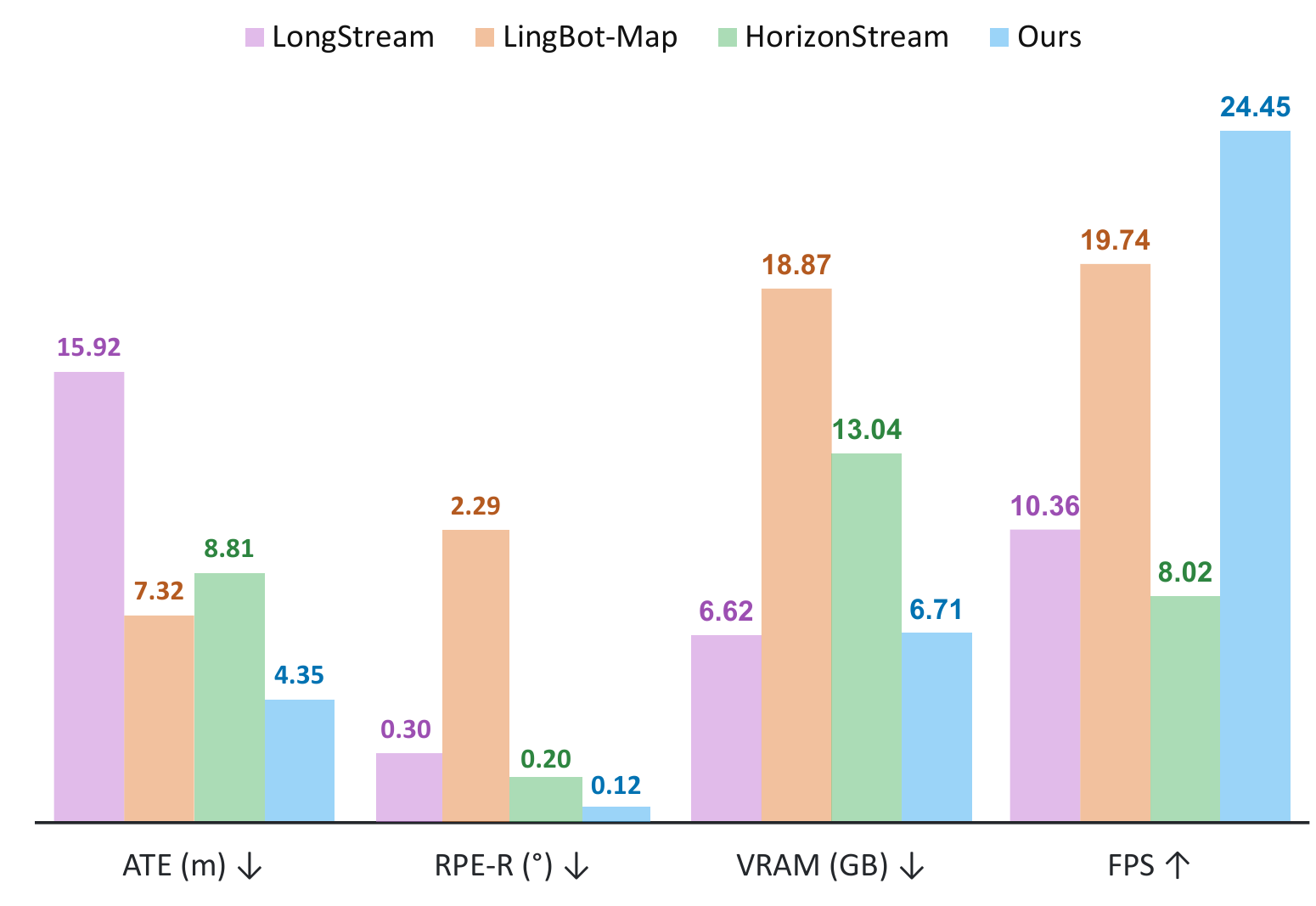}

  \vspace{-0.em}
  \caption{
    Pose accuracy on Oxford Spires and streaming inference efficiency on KITTI-02, measured by ATE, $\mathrm{RPE}_r$, FPS, and peak GPU memory excluding input storage, on an NVIDIA H100 GPU.
  }
  \label{fig:pose_efficiency_teaser}

  \vspace{-0.3\baselineskip}
\end{wrapfigure}





Recovering camera motion and dense 3D scene geometry from a video stream is a
fundamental capability for robotics, autonomous driving, and embodied
intelligence. Recent feed-forward reconstruction
models~\cite{dust3r,mast3r,vggt,pi3} have shown that point maps and camera
parameters can be inferred directly from unposed images using learned geometric
priors. Most of these models, however, operate on a fixed collection of views
through pairwise or global interactions. Ultra-long video streams change the
nature of the problem: observations arrive causally, memory and per-frame
computation must remain bounded, and the current view can be separated from
early reference frames by thousands of time steps and substantial camera
displacement.

\begin{figure}[!h]
  \centering
  \includegraphics[width=\textwidth]{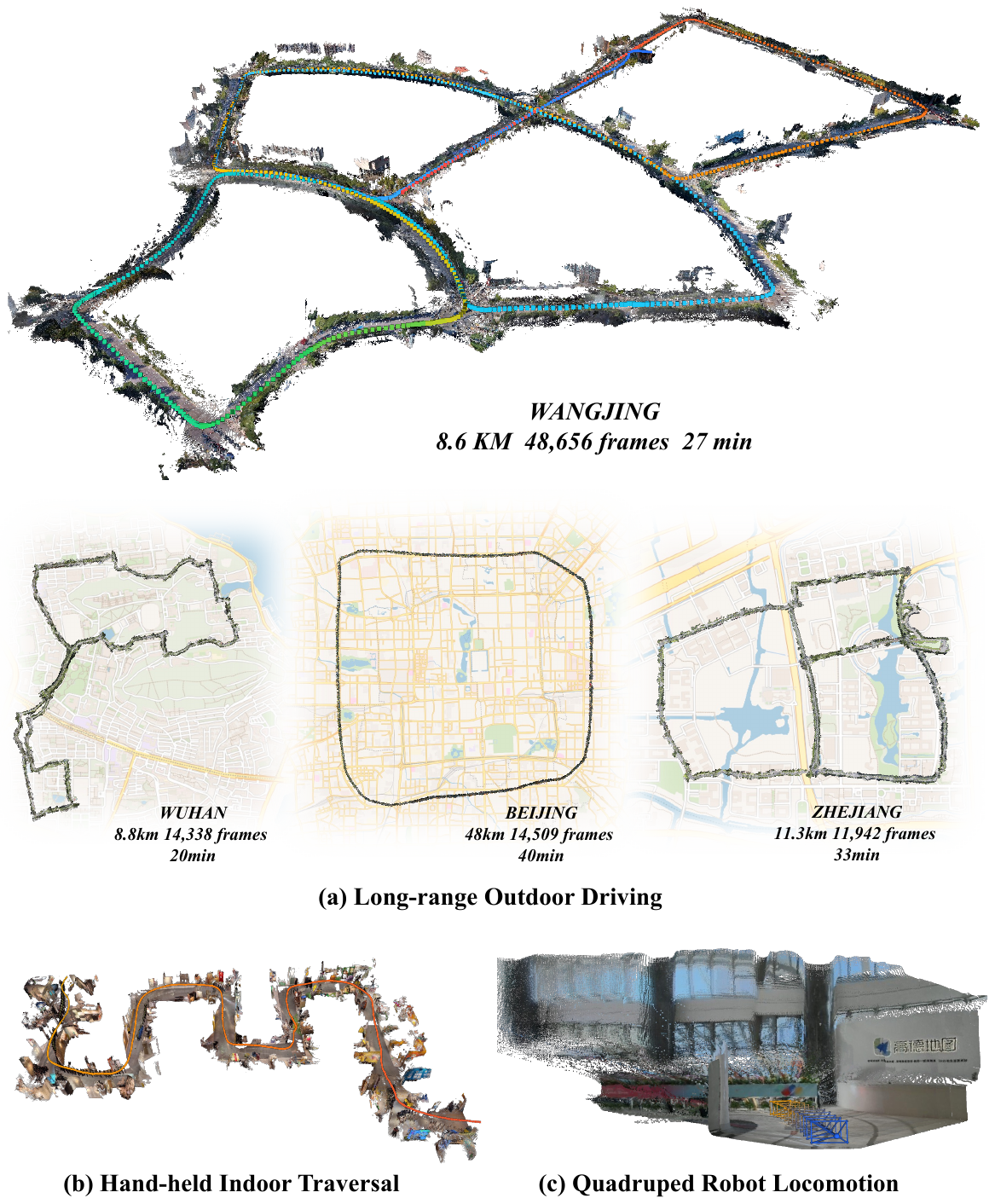}
  \caption{Results with optional loop
  closure. Our method handles diverse platforms and environments, including
  large-scale outdoor driving, hand-held indoor traversal, and quadruped robot
  locomotion.}
  \label{fig:teaser}
  \vspace{-0.3\baselineskip}
\end{figure}

Existing streaming reconstruction models adapt learned geometric priors to
online inference by propagating historical information through recurrent
representations, cached features, or explicit spatial
memories~\cite{spann3r,cut3r,streamvggt,stream3r,wint3r,slam3r,long3r,longstream,stac,ttt3r,mem3r}.
To sustain reconstruction over increasingly long sequences, these methods
employ various mechanisms to retain, compress, select, and update historical
context. More recently, LingBot-Map~\cite{lingbotmap} and
HorizonStream~\cite{horizonstream} have pushed streaming reconstruction beyond
10,000 frames by coordinating short-range observations with persistent
long-range information. These advances demonstrate the effectiveness of
long-range context, while making the management and fusion of historical state
a central part of the architecture.

We focus on a different aspect of the problem: how geometric predictions are
parameterized over time. When pose and geometry targets are defined relative to
a temporally distant reference, their estimation range grows with the sequence,
making the prediction problem increasingly difficult. We instead keep the
learned temporal context strictly local and formulate prediction targets in
local reference frames over fixed temporal ranges. This formulation brings two
benefits: it removes the need for elaborate long-range cache management, and it
keeps the per-step prediction problem unchanged as the stream grows. The model
can therefore be trained on short sequences and applied to ultra-long streams,
while the global trajectory and reconstruction are assembled incrementally
from composable local measurements.

Based on this principle, we introduce \method{}, a streamlined streaming
reconstruction model whose learned components operate within a 12-frame
temporal horizon. For each incoming frame, the model attends only to KV
features cached from the preceding 11 frames and maintains no persistent
learned long-range state. \method{} predicts a point map $P_i$ in the current
camera coordinate system, a confidence map $S_i$, and an adjacent-frame transformation
$T_{i-1\leftarrow i}$. Its local predictions transform consistently under
changes of reference frame, avoiding dependence on a fixed global anchor.
Global camera poses are recovered by chaining the adjacent transformations, and
the local point maps are transformed accordingly into the global
reconstruction. This design keeps model memory and per-frame computation
independent of the total sequence length.

However, the core challenge is error accumulation: small errors in adjacent-frame
pose estimates, particularly rotations, can compound into substantial
trajectory drift over long sequences. We address this problem at both the
estimation and supervision levels. A lightweight temporal rotation refiner
combines recent motion evolution with dense visual evidence to improve relative
rotation estimates. A composition-aware pose objective supervises
transformations over multiple temporal spans, optimizing local predictions
according to their accumulated effect rather than treating each adjacent pair
in isolation. Together, these components reduce long-horizon drift without
extending temporal context.

Extensive experiments show that \method{} generalizes from short training
sequences to ultra-long video streams and delivers superior camera tracking and
3D reconstruction performance on challenging long-sequence benchmarks,
including VBR~\cite{brizi2024vbrvisionbenchmarkrome}, KITTI~\cite{kitti}, Oxford Spires~\cite{oxfordspires}, and others. On Oxford
Spires, \method{} reduces both ATE and RPE-R by approximately 40\% relative to
the respective best prior results. Since \method{} produces a stream of local
geometric measurements, it can be readily coupled with standard SLAM modules at
inference time without retraining. Adding a loop-closure backend further
reduces the ATE to 4.02 m. These results demonstrate that a learned model with
only local temporal context can recover accurate and stable trajectories over
extended spatial and temporal horizons.

Our contributions are summarized as follows:
\begin{itemize}
    \item We establish a local-context formulation for ultra-long streaming 3D
    reconstruction, in which prediction targets remain local and their
    estimation ranges remain fixed as the sequence grows.

    \item We introduce \method{}, a streamlined streaming model that operates
    within a 12-frame temporal horizon, consisting of the current frame and KV
    features from the preceding 11 frames, and recovers global poses and
    geometry from reference-frame-consistent local predictions.

    \item We stabilize long-horizon pose composition through a lightweight
    temporal rotation refiner and composition-aware pose supervision,
    achieving superior performance on challenging long-sequence benchmarks
    without persistent learned long-range memory.
\end{itemize}

\section{Related Work}
\label{sec:related_work}

\subsection{Feed-Forward Visual Geometry Reconstruction}

Classical structure-from-motion (SfM) and visual SLAM systems~\cite{colmap,orbslam3} recover camera
motion and scene structure through carefully engineered pipelines that combine
correspondence estimation, geometric verification, triangulation, bundle
adjustment, and loop closure. Recent feed-forward methods
replace a substantial part of this pipeline with learned geometric priors.
DUSt3R~\cite{dust3r} introduced pointmap regression in a shared coordinate
frame, enabling two-view reconstruction without known camera intrinsics or
poses, while MASt3R~\cite{mast3r} augmented this representation with dense local
features for accurate image matching.

Subsequent methods extended this paradigm to broader reconstruction settings.
MV-DUSt3R+~\cite{mvdust3r} and Fast3R~\cite{fast3r} directly exchange
information across multiple input images, while VGGT~\cite{vggt} jointly
predicts cameras, depth, point maps, and tracks through global multi-view
interaction. Pi3~\cite{pi3} imposes permutation equivariance over the input
views, FLARE~\cite{flare} couples camera estimation with geometry and appearance
prediction, PLANA3R~\cite{plana3r} represents indoor scenes using metric planar
primitives learned through planar splatting, and MonST3R~\cite{monst3r} extends
feed-forward reconstruction to dynamic videos. These models demonstrate that
strong learned priors can replace much of the conventional reconstruction
pipeline. However, they either process a bounded collection of views jointly or
require pairwise or global alignment, and therefore do not by themselves
provide causal inference with bounded cost over an indefinitely growing video
stream.

\subsection{Streaming Reconstruction and Temporal Memory}

Streaming reconstruction propagates information from previous observations through different forms of temporal state. Spann3R~\cite{spann3r} stores geometric features in an external spatial memory, CUT3R~\cite{cut3r} maintains a compact recurrent state, and SLAM3R~\cite{slam3r} registers point maps reconstructed from overlapping local clips. Causal Transformer approaches include StreamVGGT~\cite{streamvggt} with cached temporal keys and values, and STream3R~\cite{stream3r} with decoder-only causal prediction. WinT3R~\cite{wint3r} combines sliding-window interaction with global camera tokens, while Point3R~\cite{point3r} uses an explicit spatial pointer memory.

To extend these architectures to longer sequences, subsequent methods strengthen how temporal state is retained and updated. LONG3R~\cite{long3r} introduces gated 3D spatio-temporal memory and length-increasing curriculum training. LongStream~\cite{longstream} adopts keyframe-relative prediction and periodic cache refresh to scale beyond short clips. STAC~\cite{stac} compresses cached features, TTT3R~\cite{ttt3r} applies confidence-adaptive recurrent-state updates, and Mem3R~\cite{mem3r} decouples tracking and mapping through fast-weight memory and a fixed-size token state.

LingBot-Map~\cite{lingbotmap} and HorizonStream~\cite{horizonstream} further extend reconstruction beyond 10,000 frames by explicitly coordinating short- and long-range information. LingBot-Map combines anchor context, a pose-reference window, and trajectory memory, whereas HorizonStream couples local attention with persistent linear attention. We instead retain only the KV cache of the most recent $K$ frames, without persistent or hierarchical long-range memory, and formulate outputs as local geometric measurements whose prediction range does not grow with the sequence.

\subsection{Local-to-Global Pose and Trajectory Recovery}

Separating local motion estimation from global trajectory recovery is central to visual odometry and SLAM. DROID-SLAM~\cite{droidslam} estimates dense inter-frame constraints and optimizes them over a graph, while ORB-SLAM3~\cite{orbslam3} combines keyframes, bundle adjustment, and loop closure. Learned systems adopt a similar decomposition. MASt3R-SLAM~\cite{mast3rslam} uses pointmap matching for tracking, loop closure, and global optimization, whereas SLAM3R~\cite{slam3r}, VGGT-Long~\cite{vggtlong}, and TALO~\cite{talo} construct and align local submaps over time. Anchor3R~\cite{anchor3r} integrates current-centered relative poses from transient windows through a motion graph. R3~\cite{xu2026r} improves pose estimation through relative prediction but remains tied to the first-frame reference, while LoGeR~\cite{zhang2026loger} combines chunk-wise inference with test-time global-consistency optimization. In these methods, global consistency is primarily recovered through an additional registration, optimization, or adaptation stage.

We instead incorporate local-to-global decomposition into the learned streaming predictor. The model predicts poses over bounded local prefixes rather than in a coordinate frame tied to a distant first frame, and adopts a reference-frame-equivariant architecture. A pose-chain loss supervises compositions across multiple temporal spans, while rotation refinement uses only recent visual and motion context. Both components preserve bounded memory while improving long-horizon composition. An optional, training-independent loop-closure backend can further refine the trajectory at inference time. Thus, long-horizon capability arises from local prediction and reliable composition rather than learned access to long-range memory.
\begin{figure}[!htbp]
  \centering
  \includegraphics[width=\textwidth]
  {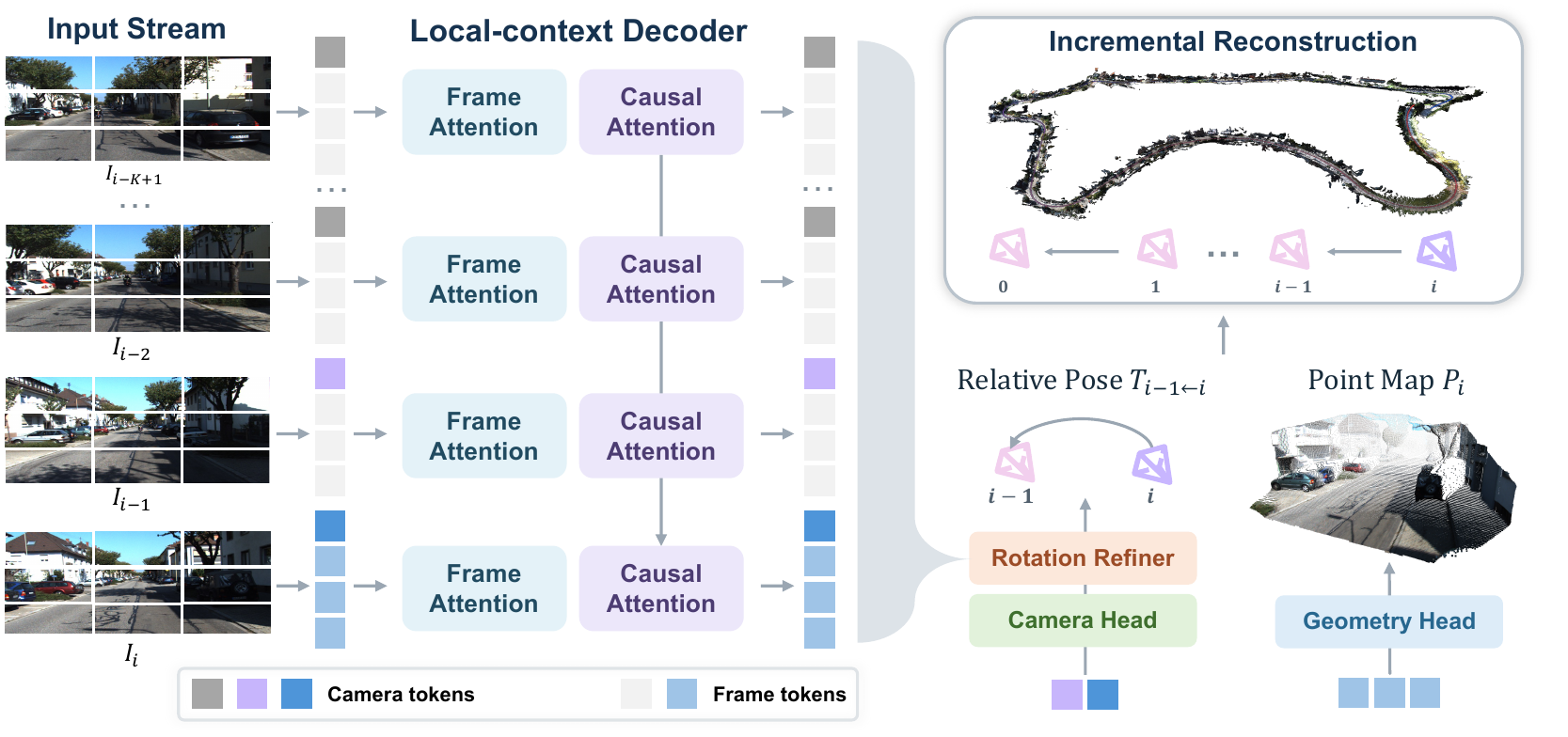}
  \caption{
  \textbf{Overview of our method.}
  For each incoming frame $I_i$, our model makes predictions using a fixed $K$-frame causal context.
  A shared image encoder followed by alternating intra-frame and windowed causal cross-frame attention predicts dense frame tokens and compact camera tokens. The geometry head predicts a local point map as well as the corresponding confidence map in the current camera coordinate system, while adjacent camera-token descriptors are used to predict the relative pose. A lightweight motion-visual rotation refiner combines the  motion and visual evidence to estimate a residual rotation and the refined adjacent transformations are composed online to recover the global camera trajectory. }
  \label{fig:overview}
\end{figure}

\section{Method}
\label{sec:method}

\subsection{Overview}

Our key design principle is to formulate long-horizon streaming reconstruction as local geometric estimation, thereby enabling a simple and scalable
architecture, as shown in Figure~\ref{fig:overview}. For each incoming frame $I_i$, \method{} predicts a point map $P_i$ in the current camera coordinate system, a corresponding confidence map $S_i$,
  and a relative pose
$T_{i-1\leftarrow i}$ to the previous frame. 
The model operates with a fixed $K$-frame temporal context, set to 12 frames in our implementation, and maintains no persistent learned long-range state. Global camera poses and scene geometry are recovered by composing these local predictions over time, keeping model memory and per-frame computation independent of sequence length.

Section~\ref{sec:local-geometry} describes the local prediction formulation and
windowed causal attention. Section~\ref{sec:refinement} introduces a lightweight
temporal refiner that improves relative rotations using recent motion and visual
context. Section~\ref{sec:loss} presents a composition-aware objective that
supervises pose chains over multiple temporal spans.



\subsection{Local Geometry Estimation}
\label{sec:local-geometry}

For each incoming frame $I_i$, our model predicts a point map $P_i$ in the current-frame coordinate system, a corresponding confidence map $S_i$, and the relative pose $T_{i-1\leftarrow i}$ of the current frame $I_i$ with respect to the previous frame $I_{i-1}$.
Our model builds on a causal-attention Transformer backbone, formulated as
\begin{equation}
\begin{aligned}
F_i
&= \operatorname{Encoder}(I_i),\\
(G_i, C_i, \mathcal{M}_i)
&= \operatorname{Decoder}([F_i,C], \mathcal{M}_{i-1}),\\
P_i
&= \operatorname{Head}_{\mathrm{point}}(G_i),\\
S_i
&= \operatorname{Head}_{\mathrm{conf}}(G_i),
\end{aligned}
\end{equation}
where $C$ denotes a set of learnable camera tokens,
$C_i=\{c_i^{\ell}\}_{\ell=1}^{L}$ denotes their decoded representations
for frame $I_i$ and $\mathcal{M}_{i-1}$ denotes the memory caches.

The relative pose is predicted directly from the camera tokens
of two adjacent frames. We first aggregate the camera tokens of the same frame into a
frame-level pose descriptor by an MLP $\phi_{\mathrm{desc}}$,
\begin{equation}
z_i
=
\frac{1}{L}
\sum_{\ell=1}^{L}
\phi_{\mathrm{desc}}
\left(c_i^{\ell}\right),
\end{equation}
and construct a pairwise descriptor~\cite{mou2016natural} between adjacent
frames
\begin{equation}
    q_{i}
    =\mathcal{R}(z_{i-1}, z_i),
    \qquad
    \mathcal{R}(x,y)=[x,y,y-x,x\odot y].
\end{equation}
The adjacent-frame relative pose is then obtained as
\begin{equation}
T_{i-1\leftarrow i}
=
\operatorname{Head}_{\mathrm{pose}}(q_{i}).
\end{equation}

The relative pose between any two frames $I_i$ and $I_j$ with $i<j$ can be
recovered by chaining the predicted adjacent-frame transformations:
\begin{equation}
\label{equ:compose}
T_{i\leftarrow j}
=
T_{i\leftarrow i+1}
T_{i+1\leftarrow i+2}
\cdots
T_{j-1\leftarrow j}
=
\prod_{k=i+1}^{j}
T_{k-1\leftarrow k}.
\end{equation}

Since geometric prediction is performed locally, we can further restrict the
temporal context to a fixed-size window.
Specifically, we replace full-history causal temporal attention with
{windowed causal temporal attention}, where frame $I_i$ attends only to
the cached keys and values of the most recent $K-1$ frames:
\begin{equation}
\mathcal{M}_{j-1}^{(K)}
=
\left\{
\mathrm{KV}_i
\right\}_{i=\max(0,j-K+1)}^{j-1}.
\end{equation}
    Cached states are discarded once they fall outside the window, resulting in $O(K)$ temporal memory and $O(NK)$ temporal-attention computation over a sequence of $N$ frames. For a fixed window size $K$, inference therefore requires constant memory while computation grows linearly with the sequence length. As this design only restricts the attention context, it remains easy to train and fully compatible with standard attention implementations, yielding a simple and scalable streaming backbone.

\begin{figure*}[t]
  \centering
  \includegraphics[width=\textwidth]
{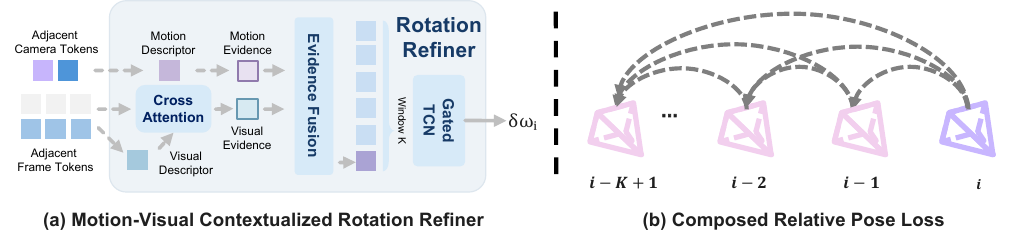}
  \caption{
Motion--visual contextualized rotation refiner with composition-aware pose supervision.  
\textbf{(a)} For each pair of adjacent frames, motion evidence is encoded from the camera tokens, while a visual descriptor queries the corresponding dense frame tokens to extract visual evidence. The two evidence streams are fused into a unified pairwise representation. These pairwise representations are then aggregated within a causal temporal window by a lightweight gated TCN, which predicts a rotation residual to refine the initial relative rotation.  
\textbf{(b)} Rather than supervising adjacent transformations independently, we compose the refined relative poses over multiple temporal gaps and compare each composed transformation with its ground-truth counterpart. This composition-aware supervision directly exposes accumulated pose-chain errors during training.}
  \label{fig:pic2}
\end{figure*}

\subsection{Rotation Refiner}
\label{sec:refinement}

Although predicting relative poses between adjacent frames avoids direct extrapolation in the global coordinate system, it can still be unreliable under rapid camera motion or challenging visual conditions. Moreover, it fails to enforce temporal consistency in the motion estimates, allowing errors to accumulate and causing long-term drift during recursive pose composition. To mitigate this issue, we introduce a Motion--Visual Contextualized Rotation Refiner that leverages recent motion history to refine unreliable relative pose estimates before trajectory composition. Since adjacent-frame translation is generally more constrained and stable than rotation, we preserve the predicted translation and apply temporal refinement only to the rotational component.

As shown in Figure~\ref{fig:pic2}(a), for each pair of adjacent frames, we construct a unified motion--visual evidence representation by integrating compact inter-frame motion evidence with spatially resolved visual evidence. Specifically, the motion evidence $f_m$ is obtained by encoding the predicted pairwise motion descriptor $q_i$ using an MLP $\phi_m$:
\begin{equation}
    f_m = \phi_m(q_i).
\end{equation}
Here, $\phi_m$, $\phi_v$, $\phi_{\mathrm{fuse}}$, and $\phi_o$ denote the MLPs used for motion encoding, visual-query encoding, motion--visual fusion, and rotation-residual prediction, respectively.

To complement the global motion evidence, we extract pair-specific visual
evidence in a coarse-to-fine manner. We first average-pool the dense patch
tokens of each frame to obtain $\bar{G}_{i-1}$ and $\bar{G}_i$, which are then
used to construct a compact visual descriptor. The resulting coarse descriptor
is used as a query to retrieve fine-grained information from the original
dense frame tokens:
\begin{equation}
    f_v
    =
    \operatorname{CrossAttn}\!\left(
        \phi_v\!\left(
            \mathcal{R}(\bar{G}_{i-1}, \bar{G}_i)
        \right),
        [G_{i-1}; G_i]
    \right),
\end{equation}
where $\mathcal{R}(\cdot,\cdot)$ denotes the operation of
pairwise descriptor. The resulting
relational descriptor provides a coarse characterization of the visual variation
between the two frames, while cross-attention~\cite{vaswani2017attention}
retrieves the corresponding spatially resolved information from the dense
features, yielding the pair-conditioned visual evidence $f_v$.

Finally, the motion and visual features are fused to yield a compact pairwise
representation $f_i$ that jointly captures the predicted camera motion and its
supporting visual context:
\begin{equation}
    f_i
    =
    \phi_{\mathrm{fuse}}\!\left(
        [f_m;f_v]
    \right).
\end{equation}

We then jointly model the current pairwise feature and its $K-1$ immediately
preceding features, forming a causal temporal window of length $K$. A
lightweight gated temporal convolutional network (TCN)~\cite{lea2017temporal}
is used to aggregate the recent motion--visual context:
\begin{equation}
\begin{aligned}
    \mathcal{W}_i
    &=
    [f_{i-K+1}, \ldots, f_i],\\
    \delta\omega_i
    &=
    \phi_o\!\left(
        \mathcal{T}_h(\mathcal{W}_i)
        \odot
        \sigma\!\left(
            \mathcal{T}_g(\mathcal{W}_i)
        \right)
    \right).
\end{aligned}
\end{equation}
Here, $\mathcal{T}_h$ and $\mathcal{T}_g$ denote two causal TCN branches
responsible for temporal feature aggregation and adaptive gating,
respectively. By contextualizing the current pairwise estimate with recent
motion--visual dynamics, the refiner predicts a residual that compensates for
unreliable local rotation estimates while preserving causality for online
inference.

The resulting axis--angle residual
$\delta\omega_i \in \mathbb{R}^3$ is mapped to $\mathrm{SO}(3)$ via the
exponential map and composed with the initial relative rotation:
\begin{equation}
    \widehat{R}_{i-1\leftarrow i}
    =
    \widetilde{R}_{i-1\leftarrow i}
    \operatorname{Exp}([\delta\omega_i]_{\times}).
\end{equation}
Guided by the temporal motion--visual context, this residual update corrects
unreliable relative-rotation estimates. The refined relative poses are then composed to recover the
camera trajectory.

\subsection{Composition-aware Loss Functions}
\label{sec:loss}

To promote consistency under recursive pose composition and mitigate error
accumulation, we introduce multi-gap relative-pose supervision. Rather than
constraining only adjacent-frame predictions, it supervises transformations
composed across multiple temporal gaps. Specifically, for a sequence of $N$
frames, we define the set of supervised frame pairs as
\begin{equation}
    \mathcal{P}
    =
    \left\{
        (i,j)
        \,\middle|\,
        0 \leq i < j < N,\;
        j-i \leq K - 1
    \right\},
\end{equation}
where $K$ denotes the window size. For each
$(i,j) \in \mathcal{P}$, we supervise the composed transformation against the
corresponding ground-truth transformation, as illustrated in
Figure~\ref{fig:pic2}(b).
To emphasize errors accumulated over longer composition chains, we assign each frame pair a weight proportional to its temporal gap and normalize the weights over all supervised pairs:
\begin{equation}
    \alpha_{ij}
    =
    \frac{(j-i)^{\gamma}}
    {\frac{1}{|\mathcal{P}|}
    \sum_{(m,n)\in\mathcal{P}}(n-m)^{\gamma}},
    \qquad 0<\gamma<1.
\end{equation}
Following $\pi^3$ \cite{pi3}, we use the same translation and rotation losses for each
composed relative pose:
\begin{equation}
    \mathcal{L}_{\mathrm{pose}}
    =
    \frac{1}{|\mathcal{P}|}
    \sum_{(i,j)\in\mathcal{P}}
    \left(
        \lambda_{\mathrm{trans}}
        \ell_{\mathrm{trans}}^{(i,j)}
        +
        \alpha_{ij}\lambda_{\mathrm{rot}}
        \ell_{\mathrm{rot}}^{(i,j)}
    \right).
    \label{eq:pose_loss}
\end{equation}

We additionally introduce a residual regularization term,
$\mathcal{L}_{\mathrm{smooth}}$, that penalizes the magnitude and temporal
variation of the predicted rotation residuals, thereby encouraging
small-magnitude and temporally consistent refinements.

For local geometric supervision, we retain the point-map loss
$\mathcal{L}_{\mathrm{pts}}$, surface-normal loss
$\mathcal{L}_{\mathrm{normal}}$, and confidence loss
$\mathcal{L}_{\mathrm{conf}}$~\cite{pi3}, which supervise 3D structure,
surface orientation, and pointwise prediction reliability, respectively.

The overall training objective is
\begin{equation}
    \mathcal{L}
    =
    \lambda_{\mathrm{pose}}\mathcal{L}_{\mathrm{pose}}
    +
    \lambda_{\mathrm{smooth}}\mathcal{L}_{\mathrm{smooth}}
    +
    \lambda_{\mathrm{pts}}\mathcal{L}_{\mathrm{pts}}
    +
    \lambda_{\mathrm{normal}}\mathcal{L}_{\mathrm{normal}}
    +
    \lambda_{\mathrm{conf}}\mathcal{L}_{\mathrm{conf}},
    \label{eq:total_loss}
\end{equation}
where the $\lambda$ coefficients balance the corresponding supervision
terms.

\section{Experiments}
\label{sec:experiments}

\subsection{Experimental Setup}
\label{sec:exp_setup}

\paragraph{Training data}
We train on a diverse mixture of 30 synthetic and real-world datasets covering
indoor scenes, outdoor environments, autonomous driving, handheld capture, and
aerial trajectories. As summarized in Table~\ref{tab:training_data}, synthetic
and real-world data account for 62.05\% and 37.95\% of the sampling
distribution, respectively. We assign larger sampling probabilities to datasets containing long, temporally coherent trajectories, while sampling from the remaining datasets to preserve diversity in scene appearance, geometry, and camera motion.

\begin{table*}[t]
  \centering
  \caption{Training-data composition. Sampling ratios are obtained by
  normalizing the dataset weights used by the training sampler. Internal data are divided into two categories: \emph{Internal Synthetic} and \emph{Internal Real}.}
  \label{tab:training_data}
  \scriptsize
  \setlength{\tabcolsep}{6pt}
  \begin{tabular}{lr|lr}
    \toprule
    \multicolumn{2}{c|}{Synthetic data} &
    \multicolumn{2}{c}{Real-world data} \\
    Dataset & Ratio (\%) & Dataset & Ratio (\%) \\
    \midrule
    TartanAir-v2~\cite{wang2020tartanair}       & 8.96 & DL3DV~\cite{ling2024dl3dv}             & 8.96 \\
    TartanAir~\cite{wang2020tartanair}           & 6.72 & Waymo~\cite{waymo}              & 5.38 \\
    OmniWorld-Game~\cite{zhou2025omniworld}     & 8.96 & ARKitScenes~\cite{baruch2021arkitscenes}         & 1.34 \\
    TartanGround~\cite{patel2025tartanground}       & 6.72 & ScanNet++~\cite{yeshwanth2023scannet++}           & 3.81 \\
    VirtualKITTI2~\cite{vkitt2}    & 6.27 & WildRGBD~\cite{wildrgbd}            & 2.91 \\
    HyperSim~\cite{roberts2021hypersim}           & 6.27 & ScanNet~\cite{dai2017scannet}             & 2.46 \\
    PointOdyssey~\cite{zheng2023pointodyssey}       & 0.90 & MapFree~\cite{arnold2022mapfreevisualrelocalizationmetric}             & 1.12 \\
    Unreal4K~\cite{Unreal4K}           & 1.34 & ARKitScenes HR~\cite{baruch2021arkitsceneshighres}      & 2.46 \\
    Spring~\cite{mehl2023spring}             & 0.23 & UASOL~\cite{bauer2019uasol}               & 0.45 \\
    MVS-Synth~\cite{MVS-Synth}          & 0.67 & DDAD~\cite{ddad}                & 1.34 \\
    Dynamic Replica~\cite{DynamicReplica}    & 0.45 & KITTI-360~\cite{liao2022kitti360noveldatasetbenchmarks}           & 2.02 \\
    ASE~\cite{ASE}                & 2.69 &  Holo360D~\cite{ou2026holo360dlargescalerealworlddataset}            & 2.02 \\
    MidAir~\cite{fonder2019midair}             & 2.69 & Internal Real       & 3.68  \\
    MatrixCity~\cite{li2023matrixcitylargescalecitydataset}         & 1.34 \\ 
    AirZoo~\cite{airzoo}             & 1.34 &                     &      \\
    BlendedMVS~\cite{yao2020blendedmvs}         & 2.47 &                     &      \\
    Internal Synthetic & 4.03 &                     &      \\
    \midrule
    \textbf{Synthetic total} & \textbf{62.05} &
    \textbf{Real-world total} & \textbf{37.95} \\
    \bottomrule
  \end{tabular}
\end{table*}

\paragraph{Data processing and augmentation}
We convert all camera poses to a unified camera-to-world convention, normalize
dataset-specific depth and translation units, and discard samples with invalid
poses, missing frames, or corrupted geometry. For video datasets, frames are sampled in their original temporal order using dataset-specific intervals.  
For unordered multiview datasets, we construct pseudo-sequences from a camera pose graph that favors locally coherent neighboring views.
We apply focal-length and crop perturbations and preserve the source field of view by resizing to the target width followed by center cropping or mean-color padding. For datasets with
notably off-center cameras, we retain the original non-central principal point with probability
0.9 instead of always applying principal-point-centered cropping. Photometric
augmentation includes brightness, contrast, saturation, hue, and gamma jitter,
as well as JPEG degradation and image blur. Dataset-specific temporal intervals
and augmentation parameters are provided in the supplementary material.

\paragraph{Training stages}
Training consists of two main stages with a progressively extended temporal
horizon, followed by a short confidence-calibration phase.
We initialize Stage~I from the publicly released $\pi^3$ weight~\cite{pi3}. We replace the original absolute camera-pose output
with our adjacent-frame relative pose parameterization and adapt the model to causal
streaming inference, without introducing the confidence branch. Stage~I uses 32-frame clips and is trained for 32K iterations with a batch size of 48 on 48 NVIDIA H20 GPUs.
This stage establishes stable local geometry and adjacent
motion estimation.

In Stage~II, we initialize from the Stage~I checkpoint and extend the training clips from 32 to 128 frames, while keeping the local prediction window fixed at 12 frames. Longer clips provide denser temporally continuous pose supervision and expose the model to more complete pose-composition chains within each training sample. We also enable the motion--visual contextualized rotation refiner module, which refines local rotation estimates before they
are composed into the trajectory. Stage~II is trained for another 38K
iterations with a batch size of 32 on 32 AMD MI308 GPUs.

Finally, we introduce the confidence prediction branch and fine-tune it for an
additional 4K iterations. During this phase, the reconstruction and pose
estimation networks are kept fixed, and only the confidence branch is optimized. This confidence-only fine-tuning allows the
model to estimate the reliability of its dense predictions without altering
the geometry and camera trajectories learned in the preceding stages.

\begin{table}[t]
  \centering
  \caption{Evaluation datasets and statistics.}
  \label{tab:datasets}
  \scriptsize
  \setlength{\tabcolsep}{3.5pt}
  \begin{tabular}{llccc}
    \toprule
    Dataset & Task & Frames/sequence & \#Seq. & Inference stride \\
    \midrule
    KITTI~\cite{kitti} & Camera pose & 271--4,661 & 11 & 1 \\
    VBR~\cite{brizi2024vbrvisionbenchmarkrome} & Camera pose & 8,815--18,846 & 7 & 1 \\
    Oxford Spires~\cite{oxfordspires} & Pose / Reconstruction & 3,821--3,840 & 10 & 1 \\
    7Scenes~\cite{7scene} & Reconstruction & 500--1,000 & 7 & 1 \\
    TUM-Dynamic~\cite{tum} & Reconstruction & 707--1,261 & 8 & 1 \\
    \bottomrule
  \end{tabular}
\end{table}

\paragraph{Implementation details}
We use the AdamW optimizer~\cite{adaw} throughout training. In Stage~I, the
learning rate reaches a peak of $1\times10^{-4}$ and is gradually annealed to
$1\times10^{-6}$. In Stage~II, we adopt differential learning rates: the newly
introduced rotation refiner module is optimized with a peak learning rate
of $5\times10^{-5}$, while the remaining model parameters are fine-tuned with
a smaller peak learning rate of $2\times10^{-5}$. This setting allows the new
module to adapt effectively while limiting disruption to the local pose and
geometry representations learned in Stage~I. During the final
confidence-calibration phase, only the confidence branch
is optimized, using a peak learning rate of $2\times10^{-5}$.

We maintain an exponential moving average (EMA)~\cite{ema} with a decay
coefficient of $0.999$ throughout training. Parameter
averaging smooths the optimization trajectory and reduces sensitivity to
transient noisy or unusually large updates, which is particularly useful for
our heterogeneous training mixture. We use a per-GPU batch size of one
sequence, bfloat16 mixed precision, and gradient clipping at 1.0.

Input images are resized to a width of 504 pixels while preserving their
aspect ratio. The resulting images are then center-cropped or mean-padded
vertically to obtain an input resolution of $504\times280$. The same causal
inference rule is applied to all test sequences.

\newcommand{\posehead}[3]{\shortstack[c]{%
  {{\fontsize{4.5pt}{5pt}\selectfont\bfseries\strut #1}}\\
  {\tiny\strut\textcolor{gray}{#2f}}\\
  {\tiny\strut\textcolor{blue!65!black}{#3 km}}}}
\newcommand{\metrichead}[2]{%
  \raisebox{4pt}{\shortstack[c]{%
    {\fontsize{5pt}{5.6pt}\selectfont\strut #1}\\[-0.2pt]
    {\fontsize{5pt}{5.6pt}\selectfont\strut #2}}}}
\newcommand{\posebest}[1]{{\bfseries\color[rgb]{0.12,0.42,0.70}#1}}
\newcommand{\posesecond}[1]{\underline{#1}}

\begin{table*}[t]
  \centering
  \caption{Camera pose estimation results on KITTI. Lower is better for all metrics.
  Within streaming methods, best and second-best results are marked in blue and
  underlined, respectively. $^*$ denotes methods that require camera
  intrinsics, and $\ddagger$ denotes results quoted from the corresponding
  paper rather than reproduced under our evaluation pipeline. w/ LC indicates
  results with loop closure enabled.}
  \label{tab:kitti_pose}
  \scriptsize
  \setlength{\tabcolsep}{2.2pt}
  \renewcommand{\arraystretch}{1.16}
  \resizebox{\textwidth}{!}{%
  \begin{tabular}{ll*{11}{c}|ccc}
    \toprule
    & & \multicolumn{11}{c|}{{\scriptsize Per-sequence ATE (m) $\downarrow$}} &
    \multicolumn{3}{c}{{\scriptsize Sequence averages}} \\
    \cmidrule(lr){3-13}\cmidrule(l){14-16}
    Group & Method &
    \posehead{00}{4541}{3.70} &
    \posehead{01}{1101}{2.50} &
    \posehead{02}{4661}{5.10} &
    \posehead{03}{801}{0.60} &
    \posehead{04}{271}{0.40} &
    \posehead{05}{2761}{2.20} &
    \posehead{06}{1101}{1.20} &
    \posehead{07}{1101}{0.70} &
    \posehead{08}{4071}{3.20} &
    \posehead{09}{1591}{1.70} &
    \posehead{10}{1201}{0.90} &
    \metrichead{Avg.}{ATE (m)$\downarrow$} &
    \metrichead{Avg.}{$\mathrm{RPE}_r$ ($^\circ$)$\downarrow$} &
    \metrichead{Avg.}{$\mathrm{RPE}_t$ (m)$\downarrow$} \\
    \midrule

    \multirow{8}{*}{\rotatebox[origin=c]{90}{Opt.-based}}
    & VGGT-SLAM
    & 27.99 & 214.23 & 260.88 & 53.50 & 26.02 & 99.61
    & 23.62 & 64.52 & 243.87 & 169.10 & 80.13
    & 114.86 & 1.57 & 2.73 \\

    & MASt3R-SLAM
    & 190.12 & 507.15 & 232.17 & 169.82 & 88.98 & 126.37
    & 133.33 & 62.92 & 199.70 & 178.03 & 151.69
    & 185.48 & 0.98 & 2.58 \\

    & VGGT-Long
    & 8.63 & 58.37 & 52.79 & 8.83 & 4.23 & 9.83
    & 4.65 & 2.67 & 73.12 & 31.84 & 27.67
    & 25.69 & 0.26 & 0.42 \\

    & SLAMFormer-$\infty$$^\ddagger$
    & 15.39 & 96.58 & 28.81 & 4.97 & 4.16 & 11.36
    & 13.54 & 8.53 & 37.00 & 18.85 & 13.93
    & 23.01 & -- & -- \\

    & Droid-SLAM$^*$
    & 131.25 & 267.79 & 217.63 & 3.40 & 1.47 & 80.41
    & 64.20 & 18.19 & 167.80 & 90.05 & 22.74
    & 96.81 & 0.23 & 0.60 \\

    & DPV-SLAM$^*$
    & 113.19 & 16.76 & 113.06 & 2.44 & 0.99 & 59.42
    & 53.32 & 18.90 & 110.36 & 74.60 & 13.74
    & 52.43 & 0.05 & 0.27 \\

    & DPVO$^*$
    & 113.17 & 16.56 & 113.37 & 2.45 & 0.98 & 59.59
    & 55.63 & 19.40 & 110.64 & 74.44 & 13.80
    & 52.73 & 0.05 & 0.27 \\

    & Droid-W$^*$
    & 48.26 & 207.57 & 52.80 & 1.76 & 4.65 & 25.16
    & 5.34 & 3.35 & 29.05 & 35.56 & 8.01
    & 38.32 & 0.05 & 0.36 \\

    \midrule
    \multirow{13}{*}{\rotatebox[origin=c]{90}{Streaming}}
    & InfiniteVGGT
    & 186.48 & 624.69 & 289.03 & 168.42 & 73.88 & 143.81
    & 116.65 & 85.53 & 225.14 & 214.28 & 156.87
    & 207.71 & 9.44 & 19.21 \\

    & OVGGT
    & 185.65 & 716.66 & 304.04 & 155.78 & 65.01 & 155.48
    & 108.51 & 85.92 & 246.97 & 215.64 & 178.67
    & 219.85 & 11.02 & 18.02 \\

    & Stream3R-w
    & 188.66 & 695.02 & 301.28 & 161.68 & 98.99 & 159.85
    & 122.82 & 87.19 & 264.61 & 219.90 & 194.32
    & 226.76 & 7.03 & 11.52 \\

    & CUT3R
    & 185.21 & 654.37 & 300.51 & 157.95 & 23.72 & 153.77
    & 133.09 & 68.57 & 230.95 & 206.95 & 184.34
    & 209.04 & 2.94 & 2.79 \\

    & CUT3R w/ reset
    & 190.70 & 91.56 & 269.58 & 20.95 & 5.97 & 104.88
    & 72.01 & 27.44 & 164.69 & 93.11 & 41.53
    & 98.40 & 0.55 & 0.65 \\

    & TTT3R
    & 165.12 & 524.16 & 272.38 & 105.06 & 12.32 & 146.24
    & 131.02 & 61.75 & 262.88 & 187.14 & 124.13
    & 181.11 & 3.85 & 3.49 \\

    & TTT3R w/ reset
    & 118.19 & 102.98 & 242.05 & 14.37 & 3.52 & 31.62
    & 37.66 & 11.99 & 81.52 & 79.10 & 29.81
    & 68.44 & 0.49 & 0.60 \\

    & LongStream
    & 90.02 & 63.46 & 236.42 & 5.04 & 2.57 & 80.62
    & 12.76 & 13.84 & 89.61 & 95.07 & 25.67
    & 65.01 & 0.35 & 0.33 \\

    & LingBot-Map
    & 33.57 & 151.48 & \posesecond{66.02} & \posebest{2.72}
    & \posesecond{2.09} & 12.79 & \posesecond{6.06} & 4.82
    & \posebest{13.36} & 19.50 & \posebest{8.01}
    & 29.13 & 0.61 & 3.04 \\

    & HorizonStream
    & 29.67 & \posesecond{24.15} & 97.24 & 5.53
    & \posebest{0.72} & 13.94 & \posebest{5.46} & 6.12
    & 22.63 & 29.79 & \posesecond{12.35}
    & 22.51 & \posesecond{0.17} & 0.11 \\

    & HorizonStream w/ LC
    & \posesecond{15.45} & \posesecond{24.15} & 71.01 & 5.53
    & \posebest{0.72} & \posesecond{7.84} & 7.54 & 3.14
    & 22.63 & 27.45 & \posesecond{12.35}
    & \posesecond{17.98} & 0.30 & \posebest{0.09} \\

    \cmidrule(l){2-16}
    & Ours
    & 21.86 & \posebest{21.55} & 69.87 & \posesecond{4.38}
    & 2.93 & 13.54 & 12.04 & \posesecond{2.02}
    & \posesecond{22.30} & \posesecond{14.11} & 16.17
    & 18.25 & \posebest{0.10} & \posesecond{0.10} \\

    & Ours w/ LC
    & \posebest{11.70} & \posebest{21.55} & \posebest{39.00}
    & \posesecond{4.38} & 2.93 & \posebest{6.00} & 8.70
    & \posebest{1.65} & \posesecond{22.30} & \posebest{13.99}
    & 16.17 & \posebest{13.49} & \posebest{0.10}
    & \posesecond{0.10} \\

    \bottomrule
   \end{tabular}}
\end{table*}

\begin{table*}[t]
  \centering
  \caption{Camera pose estimation on Oxford Spires. Within streaming methods,
  best and second-best results are marked in blue and underlined, respectively.
  $^*$ denotes methods that require camera intrinsics.}
  \label{tab:oxford_pose}
  \scriptsize
  \setlength{\tabcolsep}{2.2pt}
  \renewcommand{\arraystretch}{1.16}
  \resizebox{\textwidth}{!}{%
  \begin{tabular}{ll*{10}{c}|ccc}
    \toprule
    & & \multicolumn{10}{c|}{{\scriptsize Per-sequence ATE (m) $\downarrow$}} &
    \multicolumn{3}{c}{{\scriptsize Sequence averages}} \\
    \cmidrule(lr){3-12}\cmidrule(l){13-15}
    Group & Method &
    \posehead{Keble-2}{3840}{0.20} &
    \posehead{Keble-3}{3840}{0.18} &
    \posehead{Keble-4}{3840}{0.23} &
    \posehead{Keble-5}{3840}{0.21} &
    \posehead{Observ.-1}{3840}{0.27} &
    \posehead{Observ.-2}{3839}{0.28} &
    \posehead{Christ-2}{3821}{0.25} &
    \posehead{Christ-3}{3840}{0.18} &
    \posehead{Christ-5}{3840}{0.80} &
    \posehead{Bodleian-2}{3840}{0.55} &
    \metrichead{Avg.}{ATE (m)$\downarrow$} &
    \metrichead{Avg.}{$\mathrm{RPE}_r$ ($^\circ$)$\downarrow$} &
    \metrichead{Avg.}{$\mathrm{RPE}_t$ (m)$\downarrow$} \\
    \midrule

    \multirow{7}{*}{\rotatebox[origin=c]{90}{Opt.-based}}
    & VGGT-SLAM
    & 2.64 & 3.78 & 3.89 & 1.86 & 5.74 & 3.39 & 7.41 & 1.02
    & 12.94 & 18.83 & 6.15 & 0.39 & 0.05 \\

    & MASt3R-SLAM
    & 7.06 & 18.03 & 11.27 & 14.64 & 10.26 & 10.30 & 2.90 & 6.95
    & 19.85 & 19.99 & 12.12 & 0.36 & 0.12 \\

    & VGGT-Long
    & 9.07 & 9.37 & 11.77 & 3.96 & 6.03 & 4.05 & 2.41 & 0.48
    & 14.38 & 12.59 & 7.41 & 0.29 & 0.08 \\

    & Droid-SLAM$^*$
    & 16.79 & 15.99 & 29.40 & 21.00 & 0.87 & 1.08 & 1.31 & 1.15
    & 7.58 & 1.82 & 9.70 & 0.12 & 0.03 \\

    & DPV-SLAM$^*$
    & 0.20 & 1.31 & 0.03 & 0.30 & 0.16 & 0.45 & 0.15 & 0.04
    & 25.77 & 1.08 & 2.95 & 0.15 & 0.03 \\

    & DPVO$^*$
    & 0.12 & 0.84 & 0.05 & 0.29 & 0.16 & 0.51 & 0.15 & 0.35
    & 12.04 & 1.08 & 1.56 & 0.16 & 0.01 \\

    & Droid-W$^*$
    & 2.45 & 0.99 & 2.64 & 4.27 & 1.07 & 1.06 & 1.63 & 0.75
    & 1.56 & 6.28 & 2.27 & 0.08 & 0.01 \\

    \midrule
    \multirow{13}{*}{\rotatebox[origin=c]{90}{Streaming}}
    & InfiniteVGGT
    & 25.02 & 20.39 & 25.09 & 24.15 & 29.10 & 28.14 & 33.02 & 12.95
    & 40.39 & 79.32 & 31.76 & 5.30 & 2.70 \\

    & OVGGT
    & 32.97 & 23.01 & 28.53 & 23.23 & 28.51 & 27.75 & 30.01 & 13.09
    & 42.35 & 82.41 & 33.19 & 7.83 & 3.61 \\

    & Stream3R-w
    & 31.28 & 22.39 & 31.07 & 24.63 & 28.38 & 27.18 & 38.12 & 12.99
    & 42.39 & 82.66 & 34.11 & 9.05 & 2.67 \\

    & CUT3R
    & 31.68 & 21.67 & 22.76 & 25.07 & 26.77 & 29.41 & 33.93 & 13.06
    & 43.66 & 79.93 & 32.79 & 1.78 & 0.25 \\

    & CUT3R w/ reset
    & 13.10 & 11.55 & 15.76 & 8.98 & 13.58 & 15.60 & 12.36 & 5.71
    & 31.31 & 19.42 & 14.74 & 0.66 & 0.05 \\

    & TTT3R
    & 24.66 & 22.56 & 27.33 & 18.67 & 26.94 & 28.18 & 19.83 & 9.68
    & 26.76 & 55.01 & 25.96 & 3.19 & 0.31 \\

    & TTT3R w/ reset
    & 8.27 & 7.26 & 8.77 & 7.55 & 9.43 & 10.30 & 4.56 & 2.11
    & \posesecond{21.75} & 14.00 & 9.40 & 0.29 & 0.04 \\

    & LongStream
    & 11.57 & 9.44 & 16.88 & 10.67 & 12.43 & 14.34 & 17.93 & 6.42
    & 35.49 & 24.02 & 15.92 & 0.30 & 0.06 \\

    & LingBot-Map
    & \posebest{2.71} & \posesecond{3.04} & \posesecond{2.04}
    & \posebest{3.23} & \posesecond{5.38} & \posesecond{4.43}
    & \posebest{2.21} & \posesecond{0.74} & 37.99 & 11.43
    & 7.32 & 2.29 & 0.53 \\

    & HorizonStream
    & 5.79 & 6.47 & 7.36 & 6.27 & 6.33 & 5.30 & 10.69 & 3.11
    & 25.35 & \posesecond{11.41} & 8.81
    & \posesecond{0.20} & \posesecond{0.03} \\

    & HorizonStream w/ LC
    & 5.79 & 6.47 & \posebest{1.25} & 6.24 & 6.33 & 5.30 & 10.69
    & \posebest{0.47} & 25.35 & \posesecond{11.41} & 7.93
    & \posesecond{0.20} & \posesecond{0.03} \\

    \cmidrule(l){2-15}
    & Ours
    & \posesecond{4.24} & \posebest{1.70} & 5.08
    & \posesecond{4.44} & \posebest{4.60} & \posebest{4.40}
    & \posesecond{3.07} & 1.15 & \posebest{9.03}
    & \posebest{5.76} & \posesecond{4.35}
    & \posebest{0.12} & \posebest{0.02} \\

    & Ours w/ LC
    & \posesecond{4.24} & \posebest{1.70} & \posesecond{2.04}
    & \posesecond{4.44} & \posebest{4.60} & \posebest{4.40}
    & \posesecond{3.07} & 0.85 & \posebest{9.03}
    & \posebest{5.76} & \posebest{4.02}
    & \posebest{0.12} & \posebest{0.02} \\

    \bottomrule
  \end{tabular}}
\end{table*}

\begin{table*}[t]
  \centering
  \caption{Camera pose estimation on VBR. Within streaming methods, best
  and second-best results are marked in blue and underlined, respectively.
  $^*$ denotes methods that require camera intrinsics.}
  \label{tab:vbr_pose}
  \scriptsize
  \setlength{\tabcolsep}{2.5pt}
  \renewcommand{\arraystretch}{1.16}
  \resizebox{\textwidth}{!}{%
  \begin{tabular}{ll*{7}{c}|ccc}
    \toprule
    & & \multicolumn{7}{c|}{{\scriptsize Per-sequence ATE (m) $\downarrow$}} &
    \multicolumn{3}{c}{{\scriptsize Sequence averages}} \\
    \cmidrule(lr){3-9}\cmidrule(l){10-12}
    Group & Method &
    \posehead{Colosseo-0}{8815}{1.45} &
    \posehead{Campus-0}{12042}{2.73} &
    \posehead{Campus-1}{11671}{2.95} &
    \posehead{Pincio-0}{11142}{1.27} &
    \posehead{Spagna-0}{14141}{1.56} &
    \posehead{Diag-0}{10021}{1.02} &
    \posehead{Ciampino-1}{18846}{5.20} &
    \metrichead{Avg.}{ATE (m)$\downarrow$} &
    \metrichead{Avg.}{$\mathrm{RPE}_r$ ($^\circ$)$\downarrow$} &
    \metrichead{Avg.}{$\mathrm{RPE}_t$ (m)$\downarrow$} \\
    \midrule

    \multirow{7}{*}{\rotatebox[origin=c]{90}{Opt.-based}}
    & VGGT-SLAM
    & 54.30 & 122.07 & 117.15 & 32.48 & 24.86 & 33.52 & 188.30
    & 81.81 & 1.08 & 0.31 \\

    & MASt3R-SLAM
    & 88.75 & 116.21 & 114.37 & 67.11 & 54.62 & 35.67 & 171.76
    & 92.64 & 1.25 & 0.30 \\

    & VGGT-Long
    & 44.03 & 132.67 & 115.62 & 64.29 & 59.04 & 33.66 & 179.43
    & 89.82 & 0.82 & 0.37 \\

    & Droid-SLAM$^*$
    & 98.12 & 132.85 & 119.53 & 63.21 & 64.20 & 35.86 & 198.21
    & 101.71 & 0.82 & 0.23 \\

    & DPV-SLAM$^*$
    & 86.34 & 107.85 & 70.29 & 51.56 & 55.60 & 33.87 & 105.91
    & 73.06 & 0.70 & 0.16 \\

    & DPVO$^*$
    & 90.42 & 109.72 & 58.49 & 52.83 & 57.84 & 33.84 & 102.60
    & 72.25 & 1.34 & 0.15 \\

    & Droid-W$^*$
    & 50.27 & 125.02 & 96.89 & 21.96 & 22.82 & 24.59 & 171.52
    & 73.29 & 0.58 & 0.13 \\

    \midrule
    \multirow{13}{*}{\rotatebox[origin=c]{90}{Streaming}}
    & InfiniteVGGT
    & 89.89 & 136.76 & 116.63 & 77.76 & 63.35 & 33.84 & 202.09
    & 102.90 & 8.27 & 4.84 \\

    & OVGGT
    & 96.94 & 137.74 & 117.39 & 77.76 & 62.49 & 34.84 & 201.43
    & 104.08 & 11.43 & 6.07 \\

    & Stream3R-w
    & 92.51 & 138.38 & 116.38 & 77.43 & 64.57 & 35.64 & 202.14
    & 103.87 & 7.81 & 2.72 \\

    & CUT3R
    & 99.21 & 136.32 & 117.92 & 75.05 & 62.29 & 33.66 & 202.10
    & 103.79 & 1.50 & 0.37 \\

    & CUT3R w/ reset
    & 76.68 & 59.51 & 67.21 & 48.57 & 48.55 & 26.66 & 169.18
    & 70.91 & 0.75 & 0.11 \\

    & TTT3R
    & 91.11 & 129.47 & 116.59 & 74.31 & 60.96 & 34.39 & 194.75
    & 100.23 & 3.58 & 0.50 \\

    & TTT3R w/ reset
    & 73.32 & 56.51 & 52.76 & 37.39 & 35.33 & 34.97 & 175.98
    & 66.61 & 0.81 & \posesecond{0.10} \\

    & LongStream
    & 83.01 & 120.29 & 106.92 & 71.39 & 60.81 & 30.61 & 177.51
    & 92.93 & 0.71 & 0.13 \\

    & LingBot-Map
    & \posebest{11.11} & 21.45 & 12.39 & 46.30 & 35.40
    & 16.16 & 63.36 & 29.45 & 4.42 & 2.70 \\

    & HorizonStream
    & 40.89 & 27.84 & 24.28 & 26.77 & 28.17 & 25.39 & 29.83
    & 29.02 & \posebest{0.60} & \posebest{0.04} \\

    & HorizonStream w/ LC
    & 24.66 & 27.22 & \posesecond{8.72} & \posesecond{16.29}
    & \posesecond{21.33} & \posebest{7.36} & \posesecond{17.92}
    & \posesecond{17.64} & \posesecond{0.61} & \posebest{0.04} \\

    \cmidrule(l){2-12}
    & Ours
    & 48.12 & \posesecond{11.06} & 17.69 & 21.01 & 29.24
    & 28.06 & 55.82 & 30.14 & \posebest{0.60} & \posebest{0.04} \\

    & Ours w/ LC
    & \posesecond{16.70} & \posebest{8.10} & \posebest{4.30}
    & \posebest{5.16} & \posebest{17.96} & \posesecond{9.71}
    & \posebest{8.01} & \posebest{9.99}
    & \posebest{0.60} & \posebest{0.04} \\

    \bottomrule
  \end{tabular}}
\end{table*}

\begin{figure*}[t]
  \centering
  \includegraphics[width=\textwidth]
{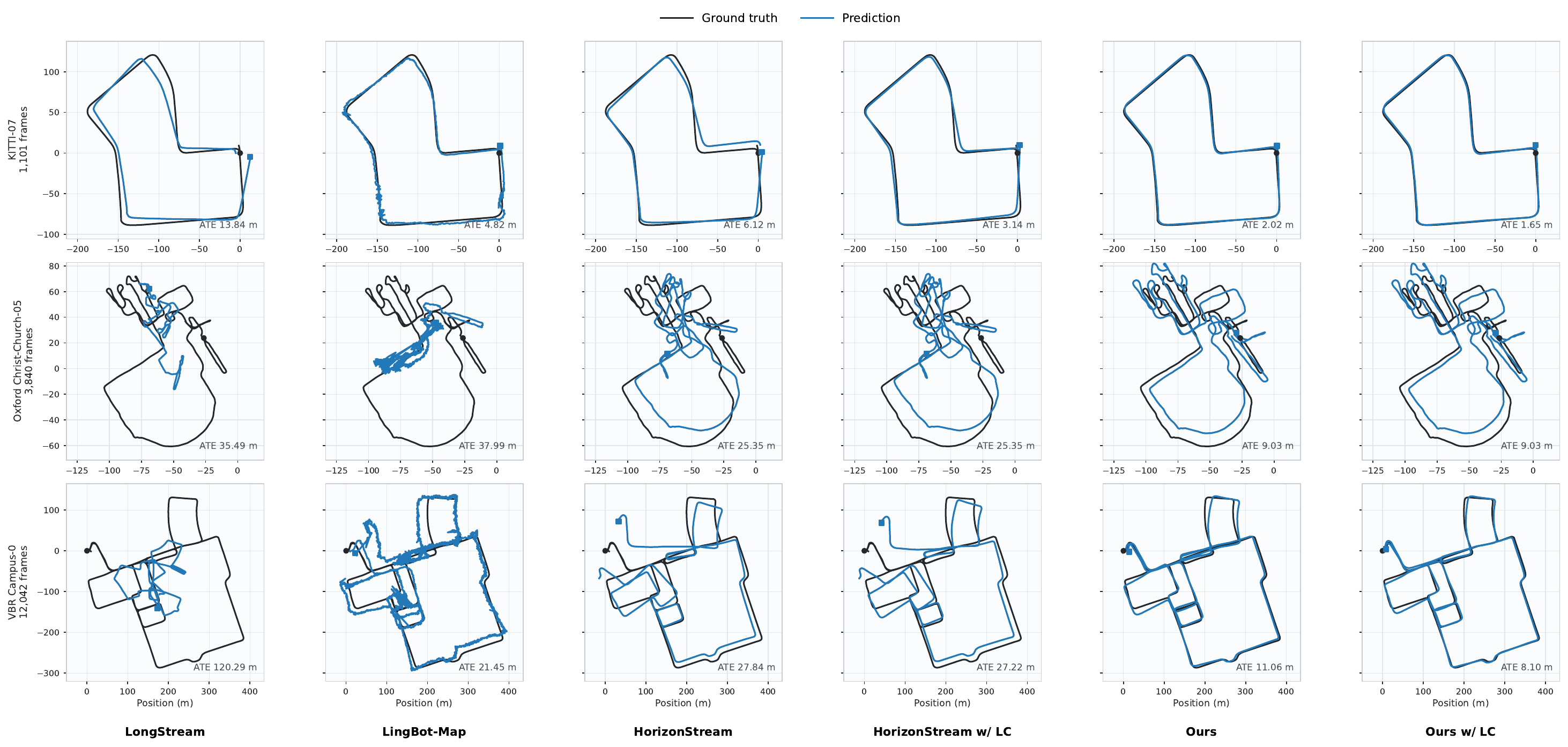}
  \caption{
  Qualitative camera pose comparison on representative sequences from
  KITTI, Oxford Spires, and VBR. 
  Predicted trajectories (blue) are aligned to the ground truth (black) using
Umeyama Sim(3) alignment, with the corresponding ATE reported in the
lower-right corner. The selected sequences cover different trajectory lengths,
scene scales, and motion patterns. Our method preserves the overall trajectory
geometry consistently across the three benchmarks, while the 
loop-closure backend further reduces accumulated global drift when revisits are detected.
  }
  \label{fig:traj_comparison}
\end{figure*}

\paragraph{Inference mode}
At inference time, the model processes the input stream causally and
continuously, without chunk-wise resets. Since its learned streaming state
consists only of a fixed-size local KV cache, inference does not require the
total number of frames to be known in advance, and the same procedure applies
regardless of sequence length. The causal KV cache is implemented using the
paged KV-cache operators in FlashInfer~\cite{ye2025flashinfer} for efficient streaming
attention. Each incoming frame is processed once to predict its local geometry
and adjacent-frame relative pose, after which the global camera trajectory and
point cloud are recovered incrementally through pose composition.

\method{} predicts relative poses from local frame windows, allowing the same
prediction interface to provide both sequential odometry constraints and
relative-pose constraints between revisited locations. The base model operates
without additional long-range memory or online optimization. When stronger
global consistency is required, we optionally attach a training-independent
SLAM-style loop-closure backend~\cite{vggtlong,horizonstream}, which refines the
accumulated trajectory without altering the underlying causal prediction
process.

To detect loop closures, revisited frame pairs are retrieved from historical
observations using FAISS-based search~\cite{johnson2019billion} over
DINOv2-SALAD descriptors~\cite{izquierdo2024optimal}. For each retrieved pair,
local windows centered on the two frames are jointly fed into \method{} to
estimate their relative pose. The sequential predictions and retrieved-pair
predictions are respectively converted into odometry and loop-closure
constraints. These constraints are incorporated into a sparse pose graph, whose
optimization refines the global trajectory.

\paragraph{Evaluation datasets}
We evaluate our method on two complementary tasks: camera pose estimation and
dense 3D reconstruction. For camera pose estimation, we use
KITTI~\cite{kitti}, Oxford Spires~\cite{oxfordspires}, and VBR~\cite{brizi2024vbrvisionbenchmarkrome}.
These benchmarks cover kilometer-scale driving sequences as well as handheld
and vehicle-mounted capture. For dense 3D reconstruction, we report results on
7Scenes~\cite{7scene}
, TUM-Dynamic~\cite{tum}, and Oxford
Spires~\cite{oxfordspires}, covering indoor and outdoor environments as well
as static and dynamic scenes. All methods process every evaluated sequence
once in temporal order with a stride of 1. For 7Scenes, we select
\texttt{seq-01} from each of its seven scenes; for all other benchmarks, we
use all sequences in the corresponding evaluation set. Dataset statistics are
summarized in Table~\ref{tab:datasets}.

\paragraph{Baseline methods}
We compare against two groups of methods. Streaming reconstruction baselines
include LingBot-Map~\cite{lingbotmap}, HorizonStream~\cite{horizonstream},
LongStream~\cite{longstream}, InfiniteVGGT~\cite{yuan2026infinitevggt}, OVGGT~\cite{lu2026ovggt},
CUT3R~\cite{cut3r}, TTT3R~\cite{ttt3r}, and Stream3R-w~\cite{stream3r}.
We additionally compare camera trajectories with optimization-based methods including VGGT-SLAM~\cite{maggio2025vggtslam}, MASt3R-SLAM~\cite{mast3rslam},
VGGT-Long~\cite{vggtlong}, SLAMFormer-$\infty$~\cite{fang2026slamformerinftyinfiniteslamtransformer}, Droid-SLAM~\cite{droidslam}, DPV-SLAM~\cite{lipson2024deeppatchvisualslam},
DPVO~\cite{dpvo}, and Droid-W~\cite{droid-W}. Droid-SLAM, DPV-SLAM, DPVO, and Droid-W use camera
intrinsics, whereas our method estimates camera motion directly from the RGB
stream without requiring test-time intrinsics. 
Since SLAMFormer-$\infty$ is not publicly available, we quote its KITTI ATE directly from the paper and all other baselines are evaluated using their publicly released
official implementations and checkpoints under our unified protocol. CUT3R and TTT3R process each complete sequence with and without state reset,
and Stream3R-w uses its official window mode with a window size of 5.

\paragraph{Evaluation protocol and metrics}
All methods process each test sequence in temporal order and all benchmarks are evaluated with a temporal stride of
1. For camera pose evaluation, each predicted trajectory is aligned to the
ground truth using Umeyama Sim(3) alignment. We report the RMSE of  Absolute Trajectory
Error (ATE), relative translational pose error ($\mathrm{RPE}_t$), and relative rotational
pose error ($\mathrm{RPE}_r$). For dense reconstruction, predicted point maps are first
resampled to the ground-truth evaluation grid using nearest-neighbor
interpolation. Following LingBot-Map~\cite{lingbotmap}, the predicted and
ground-truth point clouds are aligned using Umeyama registration, voxelized,
and refined with point-to-point ICP. We use voxel sizes of $4/512$ m for
7Scenes and TUM-Dynamic and $0.05$ m for Oxford Spires, and report Chamfer Distance (CD) and F1 at a distance threshold of $0.25$ m (7Scenes and TUM-Dynamic) and $4$
m (Oxford Spires). For Oxford Spires, inference is performed on every frame, while point-cloud
evaluation uses every 10th frame to keep evaluation tractable. Lower is better
for all metrics except F1.

\begin{figure*}[t]
  \centering

  \begin{subfigure}[t]{0.495\textwidth}
    \centering
    \includegraphics[width=\linewidth]
    {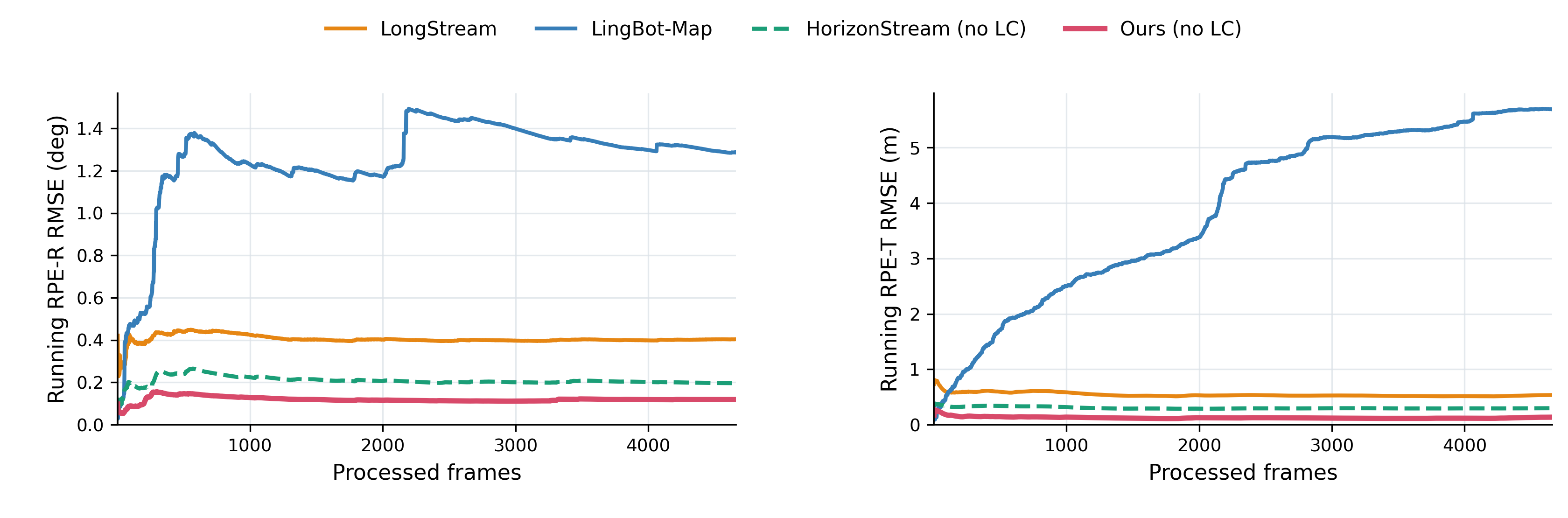}
    \caption{ }
    \label{fig:kitti_rpe_running}
  \end{subfigure}
  \hfill
  \begin{subfigure}[t]{0.495\textwidth}
    \centering
    \includegraphics[width=\linewidth]
    {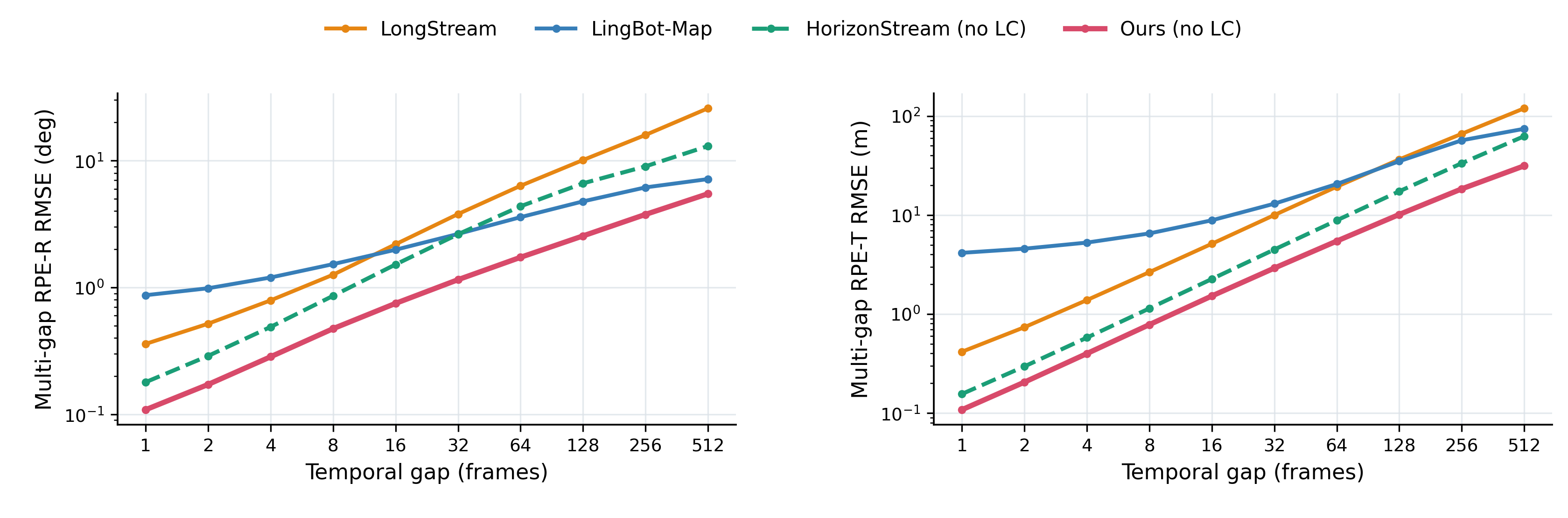}
    \caption{ }
    \label{fig:kitti_rpe_multigap}
  \end{subfigure}

  \caption{
  Long-horizon pose stability on KITTI.
  (a) Running RMSE of adjacent-frame $\mathrm{RPE}_r$ and $\mathrm{RPE}_t$ on the
  KITTI-02 sequence as the number of processed frames increases.
  (b) Multi-gap $\mathrm{RPE}_r$ and $\mathrm{RPE}_t$ averaged over all KITTI sequences
  as the temporal gap increases.
  All methods are evaluated causally without loop closure.
  }
  \label{fig:kitti_rpe_stability}
\end{figure*}

\subsection{Camera Pose Estimation}
\label{sec:pose_results}


\paragraph{Overall results}

Tables~\ref{tab:kitti_pose}, \ref{tab:oxford_pose}, and
\ref{tab:vbr_pose} report camera pose estimation results on KITTI, Oxford
Spires, and VBR, respectively. Across the three benchmarks, our method achieves strong long-horizon trajectory accuracy using only a compact local temporal context, without maintaining a persistent global reference or long-range temporal memory.

On KITTI, our method achieves the lowest average ATE among the compared
methods without and with loop closure, respectively, while also obtaining strong relative-pose accuracy. On Oxford Spires, our method achieves an average ATE of
4.35 m without loop closure, together with the best $\mathrm{RPE}_r$ and $\mathrm{RPE}_t$ among the streaming methods.
The optional loop-closure backend further reduces the average ATE to 4.02 m.
Although some optimization-based methods (DPV-SLAM, DPVO, and Droid-W) obtain lower ATE on this benchmark, these
methods require camera intrinsics, whereas ours operates solely on the RGB
stream.

On VBR, which consists of substantially longer sequences under diverse capture conditions, our method remains competitive with HorizonStream and LingBot-Map in global trajectory accuracy without loop closure, while achieving relative-pose accuracy comparable to HorizonStream. With the optional loop-closure backend, our method further improves global trajectory accuracy, achieving the lowest average ATE among the compared methods.

\paragraph{Qualitative trajectory comparison}
Figure~\ref{fig:traj_comparison} provides a complementary view of trajectory quality beyond aggregate ATE. Across the three benchmarks, our method preserves the overall trajectory geometry while producing smooth local motion. On particularly long trajectories, our relative-pose formulation exhibits less local trajectory jitter than the anchor-based LingBot-Map, while achieving a level of local stability comparable to HorizonStream despite using only a local context.

\paragraph{Long-horizon stability}
To quantitatively characterize the local trajectory stability observed above,
we first examine how adjacent-frame relative-pose errors evolve as the
processing horizon increases. Figure~\ref{fig:kitti_rpe_stability} (a) reports the
running RMSE of $\mathrm{RPE}_r$ and $\mathrm{RPE}_t$ on the 4,661-frame KITTI-02 sequence, computed over all adjacent frame pairs observed up to each processed frame. Our method maintains the lowest
running error throughout the sequence, with no progressive degradation as more
frames are processed. This behavior is consistent with the length-independent
nature of our relative-pose formulation: the regression target remains local
regardless of the total stream length.

Stable adjacent-frame predictions, however, do not necessarily guarantee
accurate long-term pose composition, as local errors can accumulate through
repeated composition. We therefore evaluate multi-gap RPE over all KITTI
sequences in Figure~\ref{fig:kitti_rpe_stability} (b). Our method achieves the
lowest average rotational and translation errors at every evaluated gap. The slower error growth at long gaps is
consistent with the role of composed-pose supervision, which directly
constrains the multi-step predictions used to assemble the streaming
trajectory.

\paragraph{Runtime and memory efficiency}

We benchmark forward throughput and peak GPU memory, excluding input storage, on a single NVIDIA H100 GPU, as reported in Table~\ref{tab:pose_efficiency}. 
The chunk-based modes of LongStream* and HorizonStream* are reported for reference and are not directly comparable to frame-by-frame streaming results.
Our method processes the stream frame by frame at 24.45 FPS using 6.71 GB of GPU memory. Under the same frame-by-frame setting, it is faster than LongStream, LingBot-Map, and HorizonStream, while using substantially less memory than LingBot-Map and HorizonStream. CUT3R and TTT3R attain slightly higher throughput with lower memory usage, but exhibit substantially larger drift over long sequences.

\begin{table*}[t]
  \centering
  \begin{minipage}[t]{0.35\textwidth}
    \centering

    \captionof{table}{
Streaming efficiency on KITTI-02 (input storage excluded). $^*$ marks default multi-frame inference and is not directly comparable to frame-by-frame results.
    }
    \label{tab:pose_efficiency}

    \vspace{1mm}

    \fontsize{6.3pt}{7.0pt}\selectfont
    \setlength{\tabcolsep}{2.2pt}
    \renewcommand{\arraystretch}{1.08}

    \begin{tabular}{@{}lcc@{}}
      \toprule

      Method
      & FPS$\uparrow$
      & Mem.$\downarrow$ \\
      \midrule

      InfiniteVGGT  & 7.30  & 15.95 \\
      OVGGT         & 11.87  & 8.05 \\
      Stream3R-w    & 12.76  & 5.26  \\
      CUT3R         & 29.62 & 3.16 \\
      TTT3R         & 25.53 & 4.65 \\
      LongStream    & 10.36 & 6.62 \\
      LongStream*    & 23.39 & 7.50 \\
      LingBot-Map   & 19.74 & 18.87 \\
      HorizonStream & 8.02 & 13.04 \\
      HorizonStream* & 29.14 & 26.62 \\

      \midrule

      Ours          & 24.45 & 6.71 \\

      \bottomrule
    \end{tabular}

  \end{minipage}%
  \hspace{0.025\textwidth}%
  \begin{minipage}[t]{0.50\textwidth}
    \centering

    \captionof{table}{
      Dense reconstruction on 7Scenes, TUM-Dynamic, and Oxford Spires.
      CD is measured in meters ($\downarrow$), and F1 in percent
      ($\uparrow$). Best and second-best results are marked in blue
      and underlined, respectively.
    }
    \label{tab:reconstruction}

    \vspace{1mm}

    \newcommand{\reconbest}[1]{%
      {\bfseries\color[rgb]{0.12,0.42,0.70}#1}}
    \newcommand{\reconsecond}[1]{%
      \underline{#1}}
    \newcommand{\reconhead}[2]{%
      \shortstack[c]{%
        #1\\[-0.2ex]
        {\scriptsize $(\tau=#2\,\mathrm{m})$}}}

    \fontsize{6.3pt}{7.0pt}\selectfont
    \setlength{\tabcolsep}{4.0pt}
    \renewcommand{\arraystretch}{1.08}

    \begin{tabular}{@{}lcccccc@{}}
      \toprule

      &
      \multicolumn{2}{c}{\reconhead{7Scenes}{0.25}}
      &
      \multicolumn{2}{c}{\reconhead{TUM-Dyn.}{0.25}}
      &
      \multicolumn{2}{c}{\reconhead{Oxford Spires}{4}} \\

      \cmidrule(lr){2-3}
      \cmidrule(lr){4-5}
      \cmidrule(l){6-7}

      Method
      & CD$\downarrow$ & F1$\uparrow$
      & CD$\downarrow$ & F1$\uparrow$
      & CD$\downarrow$ & F1$\uparrow$ \\
      \midrule

      CUT3R
      & 0.18 & 75.18
      & 0.11 & 90.68
      & 7.26 & 41.22 \\

      TTT3R
      & 0.11 & 86.98
      & \reconbest{0.06} & \reconsecond{94.45}
      & 7.77 & 40.42 \\

      Stream3R-w
      & 0.12 & 87.57
      & 0.20 & 87.40
      & 6.44 & 47.75 \\

      InfiniteVGGT
      & 0.08 & 91.96
      & 0.10 & 92.94
      & 7.38 & 47.83 \\

      OVGGT
      & 0.07 & 92.91
      & 0.09 & 93.91
      & 6.84 & 55.38 \\

      LongStream
      & \reconsecond{0.06} & 94.60
      & 0.14 & 88.29
      & 5.69 & 53.84 \\

      LingBot-Map
      & \reconbest{0.05} & \reconsecond{96.01}
      & \reconsecond{0.08} & \reconbest{95.97}
      & \reconsecond{1.68} & \reconsecond{90.58} \\

      HorizonStream
      & 0.23 & \reconbest{98.22}
      & 0.12 & 90.97
      & 2.00 & 84.69 \\

      \midrule

      Ours
      & \reconsecond{0.06} & 94.88
      & 0.11 & 92.19
      & \reconbest{1.37} & \reconbest{91.81} \\

      \bottomrule
    \end{tabular}

  \end{minipage}
\end{table*}

\subsection{Dense 3D Reconstruction}
\label{sec:reconstruction_results}

Table~\ref{tab:reconstruction} reports dense reconstruction results on
7Scenes, TUM-Dynamic, and Oxford Spires, covering compact indoor scenes,
dynamic RGB-D sequences, and larger landmark-scale environments.
Across these different settings, our method achieves competitive reconstruction quality.

On 7Scenes and TUM-Dynamic, our method remains competitive with strong streaming
baselines across both CD and F1. Although these sequences can span hundreds or
thousands of frames, they are captured within relatively compact indoor
environments, where camera observations remain spatially confined and exhibit
substantial visual overlap over time. Persistent geometric context can therefore
remain relevant throughout the stream. Nevertheless, our method retains comparable reconstruction quality across both static and dynamic indoor scenes without maintaining a persistent global anchor or long-range geometric memory.

On Oxford Spires, the sequences cover substantially larger landmark-scale
environments than the compact indoor benchmarks. Our method achieves the lowest
CD and the highest F1 among the compared methods. This result indicates
that restricting geometric prediction to local temporal context remains
effective when reconstruction spans a much larger spatial extent.

\begin{figure*}[t]
  \centering
  \includegraphics[width=\textwidth]
  {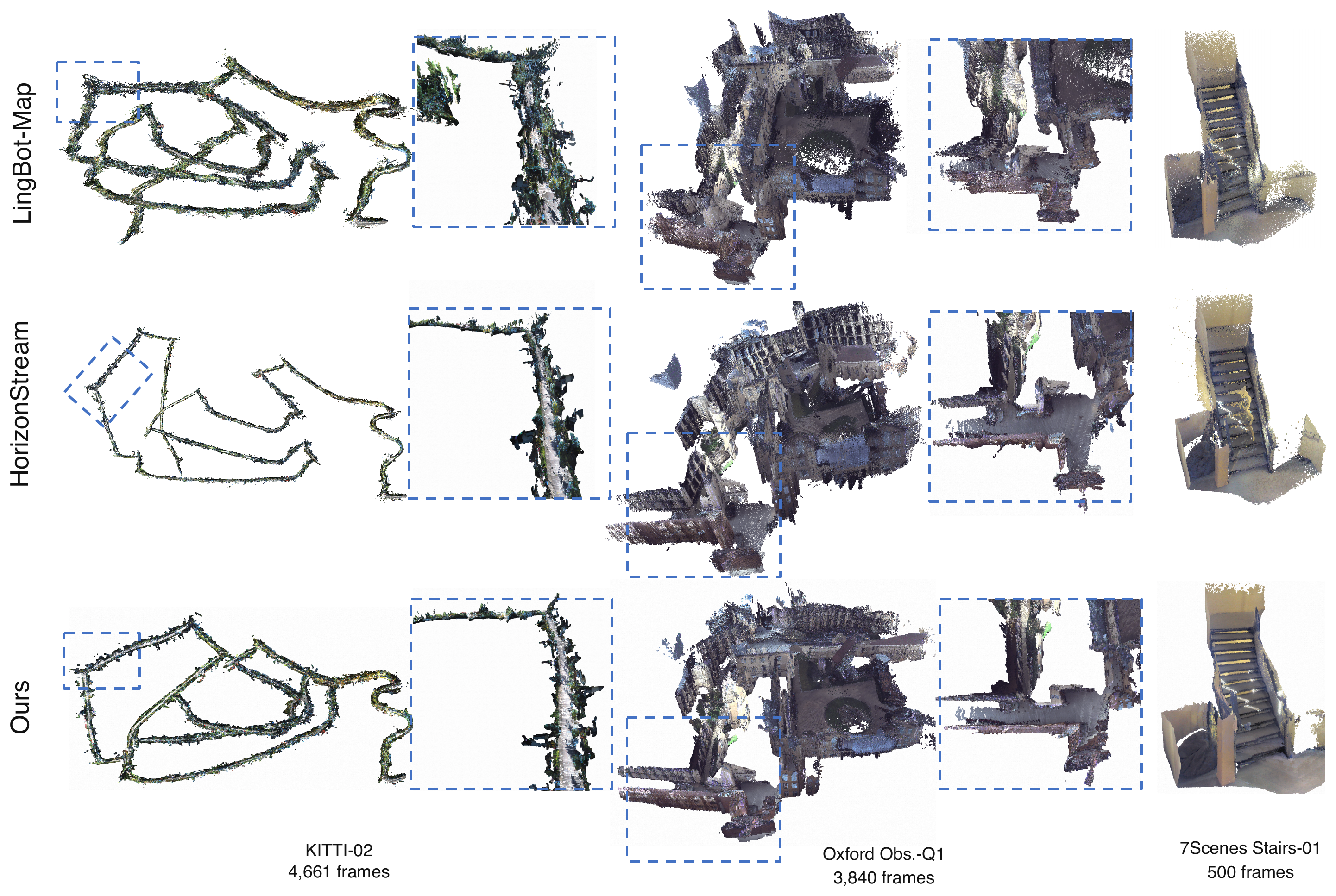}
  \caption{
  Qualitative dense reconstruction comparison on representative sequences
  from KITTI, Oxford Spires, and 7Scenes.
  Dashed boxes indicate regions shown at a larger scale for KITTI-02 and
  Oxford Obs.-Q1.
  Across different sequence lengths and scene scales, our method preserves
  coherent global structure while maintaining well-aligned local geometry
  using only local temporal context.
  }
  \label{fig:reconstruction_qualitative}
\end{figure*}

Figure~\ref{fig:reconstruction_qualitative} further compares the reconstructed
geometry on representative sequences of different lengths and scene scales.
Across KITTI-02, Oxford Obs.-Q1, and 7Scenes Stairs-01, all three methods recover
the main scene structure, while our local formulation maintains coherent
geometry from compact indoor scenes to substantially longer driving and
landmark-scale sequences. The zoomed regions on KITTI and Oxford further show
that local geometric structure remains well aligned within the global
reconstruction.

\subsection{Ablation Studies}
\label{sec:ablations}

To systematically assess the contribution of each model component, we conduct
controlled ablation studies on KITTI, Oxford, and VBR, which collectively
cover diverse motion dynamics, scene geometries, and sensing conditions. As
shown in Table~\ref{tab:ablation_all_datasets}, we begin with a baseline
trained using only adjacent-frame pose supervision and progressively introduce
the composition loss, long-sequence training, and rotation refiner.
Following the evaluation protocol described above, we report ATE,
$\mathrm{RPE}_r$, and $\mathrm{RPE}_t$.

\begin{table*}[t]
  \centering
  \caption{Ablation study on KITTI, Oxford, and VBR. The Frames column reports the number of frames in each video clip
used for training, while the Model size column reports the number of
model parameters. Results are averaged across the evaluation sequences.
Lower values indicate better performance, and the best results are
highlighted in blue.}
  \label{tab:ablation_all_datasets}

  \scriptsize
  \setlength{\tabcolsep}{2.5pt}
  \renewcommand{\arraystretch}{1.10}

    \newcommand{\reconbest}[1]{%
      {\bfseries\color[rgb]{0.12,0.42,0.70}#1}}
  \resizebox{\textwidth}{!}{%
  \begin{tabular}{lcccc|ccc|ccc|ccc}
    \toprule
    & \multicolumn{4}{c|}{Training configuration}
    & \multicolumn{3}{c|}{KITTI}
    & \multicolumn{3}{c|}{Oxford}
    & \multicolumn{3}{c}{VBR} \\
    \cmidrule(lr){2-5}
    \cmidrule(lr){6-8}
    \cmidrule(lr){9-11}
    \cmidrule(l){12-14}

    Variant
    & Frames
    & Model size
    & Refiner
    & $\mathcal{L}_{\mathrm{comp}}$
    & ATE
    & $\mathrm{RPE}_r$
    & $\mathrm{RPE}_t$
    & ATE
    & $\mathrm{RPE}_r$
    & $\mathrm{RPE}_t$
    & ATE
    & $\mathrm{RPE}_r$
    & $\mathrm{RPE}_t$ \\

    \midrule

    Baseline
    & 32 & 996M & $-$ & $-$
    & 56.60 & 0.19 & 0.17
    & 10.16 & 0.19 & 0.03
    & 52.90 & 0.83 & 0.06 \\

    + Comp. loss
    & 32 & 996M & $-$ & $\checkmark$
    & 27.66 & 0.14 & 0.13
    & 9.40  & 0.15 & 0.03
    & 44.72 & 0.63 & 0.06 \\

    + Long-seq. training
    & 128 & 996M & $-$ & $\checkmark$
    & 22.31 & 0.11 & 0.12
    & 5.95  & 0.13 & \reconbest{0.02}
    & 36.77 & 0.62 & 0.06 \\

    + Rot. Refiner
    & 128 & 1B & $\checkmark$ & $\checkmark$
    & \reconbest{18.25} & \reconbest{0.10} & \reconbest{0.10}
    & \reconbest{4.35} & \reconbest{0.12} & \reconbest{0.02}
    & \reconbest{30.14} & \reconbest{0.60} & \reconbest{0.04} \\

    \bottomrule
  \end{tabular}%
  }
\end{table*}

\paragraph{Composition-Aware Pose Supervision}
Introducing composition-aware supervision yields consistent improvements over
the 32-frame baseline across all three datasets. Although the baseline
supervises individual adjacent-frame transformations, it does not explicitly
constrain the errors that arise when these transformations are recursively
composed, leading to long-term trajectory drift. Composition-aware pose
supervision mitigates this issue by directly supervising transformations
composed across multiple adjacent steps, thereby enforcing relative-pose
consistency over different temporal gaps. These results highlight the
importance of compositional consistency for limiting long-term trajectory
drift, which pairwise pose supervision alone cannot adequately address.

\paragraph{Long-Sequence Training}
Increasing the training clip length from 32 to 128 frames further reduces both
ATE and relative-pose errors while keeping the local pose-prediction window
unchanged. Compared with short-sequence training, ATE decreases by 19.3\%,
36.7\%, and 17.8\% on KITTI, Oxford, and VBR, respectively. These gains
suggest that longer training clips provide more extensive and temporally
coherent supervision for adjacent-frame motion estimation. They also reduce
artificial clip-boundary effects and expose the model to a broader range of
local motion patterns, thereby improving the consistency of relative-pose
estimates during recursive composition.


\paragraph{Motion--Visual Contextualized Rotation Refiner}
The proposed rotation refiner adds only 4.75M parameters, representing
approximately 0.48\% of the full model, while reducing ATE by 18.2\%, 26.9\%,
and 18.0\% on KITTI, Oxford, and VBR, respectively. By jointly leveraging
complementary motion and visual cues from recent frames, the refiner predicts
more reliable rotation residuals, thereby improving relative-pose consistency
during long-horizon trajectory composition.

\section{Discussion}
\label{sec:discussion}

Our results show that the proposed local formulation is particularly effective
for long, temporally continuous streams. The model maintains stable local pose
estimation as the processing horizon grows and remains consistent under longer
pose compositions, leading to strong trajectory accuracy on KITTI, VBR, and
Oxford Spires. On compact indoor benchmarks, the gains are less pronounced and
our reconstruction results are not uniformly the best. One possible reason is
that, in spatially confined scenes with frequent revisits and high visual overlap,
persistent geometric context can remain useful over long portions of the
sequence. Our method does not explicitly retain such scene-level state, yet
remains competitive in these settings, suggesting that the local formulation
remains effective across both compact and substantially larger environments.

The optional loop-closure backend complements the local predictor when scene
revisits are available. Because it is independent of model training, loop
closure introduces sparse long-range constraints only at inference time while
leaving the learned streaming formulation unchanged. This provides additional
global correction on sequences with useful revisits, without requiring
persistent learned long-range state inside the predictor. In this sense, local
prediction and loop closure play complementary roles: the former provides a
bounded, sequence-length-independent streaming backbone, while the latter
offers optional global refinement when additional long-range constraints are
available.

The resulting modular design also leaves room for further extensions. The local
streaming backbone can incorporate advances in feed-forward reconstruction, 
caching, and inference acceleration without changing the underlying
local-to-global formulation. Future work may explore stronger handling of
dynamic scenes~\cite{chen2025easi3r,hu2026vggt4d}, external memory~\cite{wang2026amb3r, zhang2026loger}, larger-scale pretraining~\cite{wang2026vggt,lin2025depth}, and selective long-range constraints
for cases where local temporal continuity becomes ambiguous, as well as
applications to long-horizon embodied perception~\cite{agarwal2026cosmos,li2026spatial,kim2024openvla,wang2026vggt, yu20263d}, 3D-VLM~\cite{hu2026g,fan2026vlm3r,zhang2026spatialstack}, 3D Generation~\cite{qian2026gsvoxelfittingfreestructuredlatents,qian2026abotearth05generative3d,qian2026satdgen,schmid2026genreconbridginggenerativepriors,Xiang_2025_CVPR} and world modeling~\cite{sun2026vggt,huang2026gen3r3dscenegeneration,wu2026geometry}.
\section{Conclusion}
\label{sec:conclusion}



We presented \method{}, a simple streaming 3D reconstruction model whose learned predictor operates entirely within a fixed twelve-frame temporal horizon. Instead of storing or propagating persistent long-range state, \method{} predicts geometry in the current camera frame and motion between adjacent frames, keeping the estimation targets local as the sequence grows. Global camera trajectories and scene geometry are recovered by composing these local measurements over time. To make this composition reliable, a lightweight motion-visual rotation refiner improves relative rotations, while composition-aware pose supervision directly constrains their accumulated effect over multiple steps. Experiments on long driving, handheld, and landmark-scale sequences show that \method{} delivers accurate camera tracking and dense reconstruction with bounded model memory and per-frame computation. These results demonstrate that persistent learned long-range memory is not a prerequisite for ultra-long streaming reconstruction: a simple short-context model can scale effectively when its local predictions are formulated to remain accurate and reliably composable.

\section{Contributions}
\label{sec:Contributions}

\textbf{Contributor}:  Jiarong Han, Jingcheng Xiong, Yuzhou Liu, Ming Qian, Linzhe Shi, Changjie Wu\\
\textbf{Project Sponsor}:  Mu Xu, Ning Guo\\
\textbf{Project Leader}: Hang Zhang, Jiarong Han, Ming Qian

We would like to express our sincere gratitude to Hongyu Pan, Zhongxu Sun, Bentao Wang, Yuting Xu, Tianjian Ouyang, Haoming Yu, Chuzi Chen, and Zhiyang Zhang for their valuable support and contributions to this project.

\bibliographystyle{IEEEtran}
\bibliography{references_unified_updated}

\appendix

\clearpage
\appendix
\section{Data Processing}

\subsection{Temporal Sampling}
\label{app:training_data_details}

\paragraph{Video datasets}
We construct every training sample as an ordered sequence. For video datasets,
the interval $[a,b]$ denotes the range of frame-index steps between consecutive
inputs. We use three temporal sampling strategies according to the structure
and length of the source trajectory.

\begin{enumerate}
  \item \textbf{Forward fixed-stride sampling.}
  We uniformly sample one feasible interval from $[a,b]$ and use it throughout
  the clip. Frames are always traversed in chronological order. This strategy
  is used for long trajectories that can provide the requested number of
  frames without reversing direction.

  \item \textbf{Adaptive forward sampling.}
  We sample only from the forward portion of a trajectory. When the remaining
  sequence is too short for the configured maximum interval, we reduce the
  upper bound to fit the complete clip. Depending on the dataset, the resulting
  clip uses either one fixed stride or independently sampled per-step strides.

  \item \textbf{Foldback sampling.}
  We traverse the source sequence with a stride sampled from
  $[a,b]$. Upon reaching a sequence boundary, the traversal reverses direction
  and a new stride is sampled for the next segment. Foldback sampling allows a
  short source video to provide a long training clip while preserving local
  continuity and avoiding jumps between unrelated frames.

\end{enumerate}

\paragraph{Multi-view datasets}
For unordered multi-view data, we construct fixed-length sequences using a camera pose graph. Let $\theta_{ij}$ denote the relative rotation between views $i$ and $j$, and let $d_{ij}=\|\mathbf{t}_i-\mathbf{t}_j\|_2/s$ denote their translation normalized by the median nearest-neighbor camera distance $s$. For each view, we consider its $K$ nearest neighbors and retain at most $K_g$ edges satisfying $\theta_{ij}<\theta_{\max}$ and $d_{ij}<\tau_t$. Beam search then expands valid paths and keeps the best $B$ candidates according to their motion and turning costs. Repeated views are permitted only through valid graph edges, subject to a minimum unique-view ratio $\rho_{\min}$.

\begin{table}[htbp]
\centering
\small
\caption{Pose-graph sequence sampling parameters.}
\label{tab:pose_graph_sampling}
\begin{tabular}{lcccccc}
\toprule
Dataset & $\theta_{\max}$ & $\tau_t$ & $K$ & $K_g$ & $B$ & $\rho_{\min}$ \\
\midrule
HyperSim   & $20^\circ$ & $5$ & 48 & 24 & 192 & $0.50$ \\
US4K       & $20^\circ$ & $5$ & 64 & 24 & 128 & $1.00$ \\
BlendedMVS & $40^\circ$ & $5$ & 24 & 14 & 96  & $0.75$ \\
\bottomrule
\end{tabular}
\end{table}

US4K requires all views to be unique, while HyperSim and BlendedMVS allow constrained revisiting when necessary.

\paragraph{Dataset percentages}
The temporal sampling strategy or multi-view construction rule used for each
dataset is summarized in Table~\ref{tab:appendix_sampling}.

\begin{table*}[t]
  \centering
  \caption{Per-dataset sequence-construction and interval policies used for
  training, resolved from the training configuration and dataset-loader
  defaults. Interval bounds are measured in source frames. For pose-graph
  data, $b_{\mathrm{med}}$ denotes the scene's median nearest-neighbor camera
  baseline.}
  \label{tab:appendix_sampling}
  \scriptsize
  \setlength{\tabcolsep}{6pt}
  \renewcommand{\arraystretch}{1.02}
  \begin{tabular}{@{}llcl@{}}
    \toprule
    Dataset & Type & Sampling strategy & Interval / constraint \\
    \midrule
    TartanAir-v2 & Synth. & Foldback & Hard $[1,8]$, Easy $[1,16]$ \\
    TartanAir & Synth. & Adaptive forward & $[1,20]$ \\
    OmniWorld-Game & Synth. & Foldback & $[4,16]$ \\
    TartanGround & Synth. & Foldback & $[1,8]$ \\
    Virtual KITTI 2 & Synth. & Adaptive forward & $[1,5]$ \\
    HyperSim & Synth. & Pose-graph path & $\Delta R<20^\circ$, $\Delta t<5b_{\mathrm{med}}$ \\
    PointOdyssey & Synth. & Adaptive forward & $[1,4]$ \\
    Unreal4K & Synth. & Pose-graph path & $\Delta R<20^\circ$, $\Delta t<5b_{\mathrm{med}}$ \\
    Spring & Synth. & Foldback & $[1,4]$ \\
    MVS-Synth & Synth. & Foldback & $[1,4]$ \\
    Dynamic Replica & Synth. & Adaptive forward & $[1,16]$ \\
    ASE & Synth. & Foldback & $[1,2]$ \\
    MidAir & Synth. & Foldback & $[1,8]$ \\
    MatrixCity & Synth. & Forward fixed-stride & 1 \\
    AirZoo & Synth. & Foldback & $[1,8]$ \\
    BlendedMVS & Synth. & Pose-graph path & $\Delta R<40^\circ$, $\Delta t<5b_{\mathrm{med}}$ \\
    Internal Synthetic & Synth. & Forward fixed-stride & $[2,4]$ \\
    \midrule
    DL3DV & Real & Adaptive forward & $[1,20]$ \\
    Waymo & Real & Adaptive forward & $[1,8]$ \\
    ARKitScenes & Real & Foldback & 1 \\
    ScanNet++ & Real & Foldback & 1 \\
    WildRGBD & Real & Foldback & $[1,4]$ \\
    ScanNet & Real & Adaptive forward & $[1,30]$ \\
    MapFree & Real & Adaptive forward & $[1,30]$ \\
    ARKitScenes HR & Real & Foldback & 1 \\
    UASOL & Real & Adaptive forward & $[1,40]$ \\
    DDAD & Real & Forward fixed-stride & 1 \\
    KITTI-360 & Real & Forward fixed-stride & $[2,6]$ \\
    Internal Real & Real & Forward fixed-stride & $[1,2]$ \\
    Holo360D & Real & Foldback & $[1,2]$ \\
    Holo360 & Real & Foldback & $[1,2]$ \\
    \bottomrule
  \end{tabular}
\end{table*}

\subsection{Data Filtering}
We find that public datasets contain occasional annotation errors and noisy samples. By inspecting sampled training examples and analyzing the samples associated with loss spikes during training, we identify problematic data and apply dataset-specific filtering. Removing these erroneous samples substantially improves model performance.

\begin{figure*}[htbp]
    \centering
    \includegraphics[width=\textwidth]{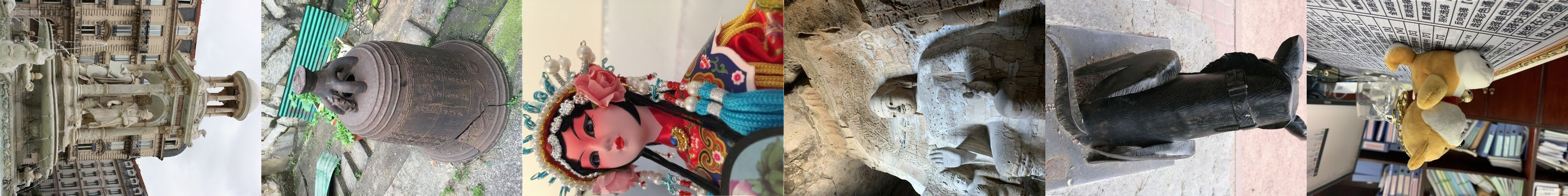}
    \caption{Examples of sideways and upside-down images in BlendedMVS.}
    \label{fig:blendedmvs_filtering}
\end{figure*}

\paragraph{BlendedMVS} 
We observed that some BlendedMVS scenes contain sideways or upside-down images, as shown in Fig.~\ref{fig:blendedmvs_filtering}. We manually identified and excluded 38 such scenes.

\begin{figure*}[htbp]
    \centering
    \includegraphics[width=0.8\textwidth]{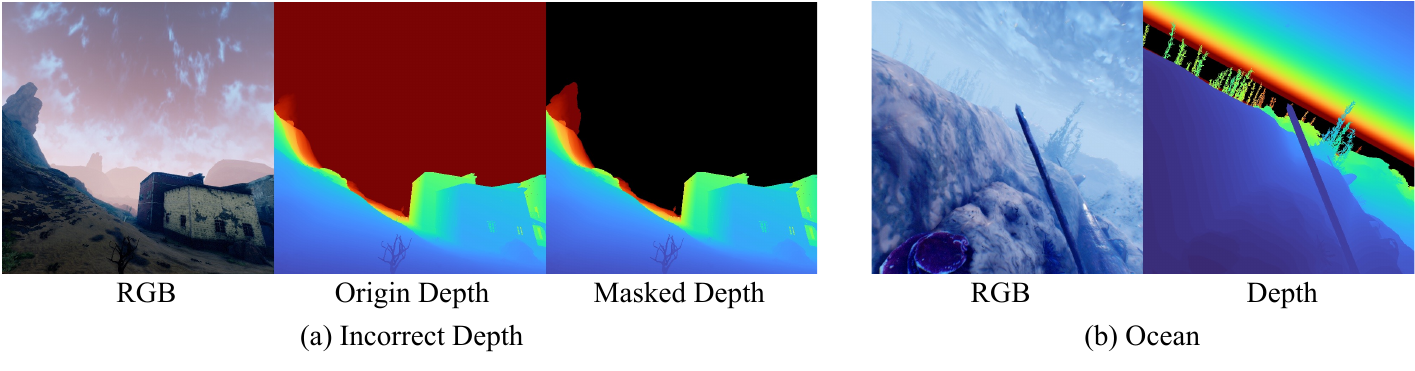}
    \caption{Examples of incorrect scenes in TartanAir and TartanAir-v2.}
    \label{fig:tartanair_filtering}
\end{figure*}

\paragraph{TartanAir \& TartanAir v2}
For the TartanAir datasets, we found that some scenes contained erroneous depth values in sky regions, as shown in Figure~\ref{fig:tartanair_filtering}(a). We invalidated the corresponding depth values using semantic masks. We also observed erroneous water-surface depths in the Ocean scene, as shown in Figure~\ref{fig:tartanair_filtering}(b), and therefore excluded this scene from both TartanAir and TartanAir-v2.

\paragraph{OmniWorld-Game}

\begin{figure}[htbp]
    \centering
    \includegraphics[width=\textwidth]{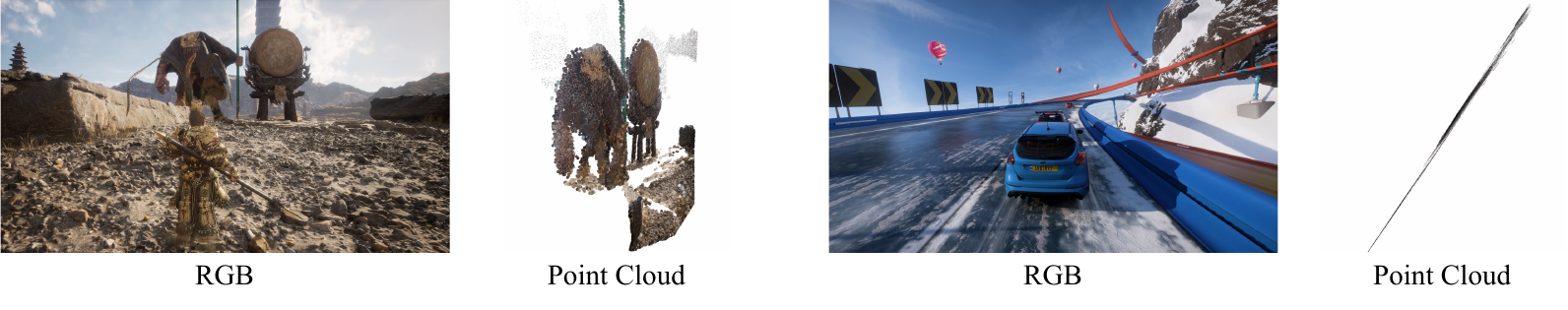}
    \caption{Examples of incorrect scenes in OmniWorld-Game.}
    \label{fig:omniworldgame_filtering}
\end{figure}

We observed erroneous geometric annotations in the dataset, particularly anomalous focal lengths in the camera intrinsics, as shown in Figure~\ref{fig:omniworldgame_filtering}. We therefore remove unreliable scenes by validating cross-frame correspondences. Specifically, we sample multiple image pairs at different temporal positions in each sequence and obtain reliable 2D correspondences using SIFT~\cite{lowe2004distinctive} feature matching and fundamental-matrix RANSAC. The matched pixels are then back-projected into 3D using the annotated depths and camera intrinsics. We estimate the relative rigid motion between the two frames and compare it with the annotated camera pose. Rotation error, translation direction and scale errors, and depth-normalized geometric residuals are jointly used to identify inconsistent frame pairs. A sequence is removed when severe inconsistencies account for at least $15\%$ of its valid frame pairs and form a cluster of at least three pairs.

\paragraph{DL3DV}

Since DL3DV is constructed from captured videos using structure-from-motion (SfM) and multi-view stereo (MVS), some scenes contain erroneous pose reconstructions. We assume that scenes with discontinuous camera trajectories are likely to suffer from reconstruction failures. Therefore, we remove any scene in which the rotation between two consecutive frames exceeds $70^\circ$, or the inter-frame translation exceeds $15 \times$ median inter-frame translation of that scene. Moreover, because the DL3DV depth maps contain noise, we first discard the farthest $2\%$ of depth values, apply a morphological opening with a $5\times 5$ kernel, and then perform erosion with a $3\times 3$ kernel to remove isolated depth outliers.

\paragraph{HyperSim}
We found that all sampled RGB frames from \texttt{ai\_003\_001/cam\_00} and \texttt{ai\_004\_009/cam\_01} were completely black; therefore, we excluded both sequences from training.

\paragraph{ScanNet++}
We observed floating artifacts within the reconstructed meshes of some scenes, which occluded large portions of otherwise valid depth. We therefore excluded the affected scenes from training.

\paragraph{ARKitScenes}
We found that the low-resolution depth maps in ARKitScenes, captured using the iPhone LiDAR sensor, are not sufficiently accurate. Therefore, we do not use these depth maps for supervision.

\subsection{Spatial Preprocessing and Augmentation}
With probability 0.9, we preserve the source field of view by resizing the
image isotropically to 504 pixels in width and producing a $504\times280$
canvas through vertical center cropping or symmetric mean-color padding. The
padding value is $(0.485,0.456,0.406)$ in normalized RGB space. Geometric
annotations and camera intrinsics are transformed consistently with the image.
Photometric transformations are applied only to valid image content; synthetic padding is restored to its
mean value afterward. The photometric pipeline comprises color jitter
(brightness, contrast, saturation, hue, and gamma), JPEG degradation, and
image blur. For 20\% of training sequences, a single set of augmentation
parameters---including the JPEG and blur decisions---is sampled and shared by
all frames. For the remaining 80\%, these parameters are sampled independently
for each frame. This mixed strategy retains temporal coherence in part of the
training data while exposing the model to frame-wise photometric variation.

\section{Runtime and Memory Evaluation Protocol}
\label{sec:app_efficiency}





We evaluate runtime and memory efficiency on the full 4,661-frame KITTI-02 sequence at stride 1 using a single NVIDIA H100 GPU. Each method uses its specified inference precision and input resolution. In addition to frame-by-frame streaming, we report the released chunk-based execution modes of LongStream and HorizonStream, denoted by an asterisk. Before measurement, we perform an untimed 64-frame warm-up and then reset the streaming state. The complete 4,661-frame sequence is subsequently processed for timing. All input frames are preloaded onto the GPU, and image loading and preprocessing are excluded. We synchronize CUDA once immediately before and after the complete forward pass; thus, the reported FPS includes model inference and all streaming-state updates, without introducing per-frame synchronization overhead.
Peak memory is reported as the maximum PyTorch-allocated GPU memory excluding the storage occupied by the preloaded input sequence. It includes model parameters, intermediate activations, cached tensors, and persistent streaming state. All methods are evaluated on the same sequence and hardware using the same timing and memory accounting protocol. The resulting throughput and memory measurements are reported in Table~\ref{tab:pose_efficiency}.

\section{More Results}
\label{sec:moreresult}

We further evaluate the generalization of our method on long videos collected outside the benchmarks in the main paper. 
These sequences are not associated with accurate ground-truth camera trajectories; therefore, the results in this
section are intended as qualitative evidence rather than quantitative
comparisons. We consider two complementary settings: low-viewpoint robot
navigation and large-scale real-world traversal.

\begin{figure*}[t]
  \centering
  \includegraphics[width=\textwidth]
  {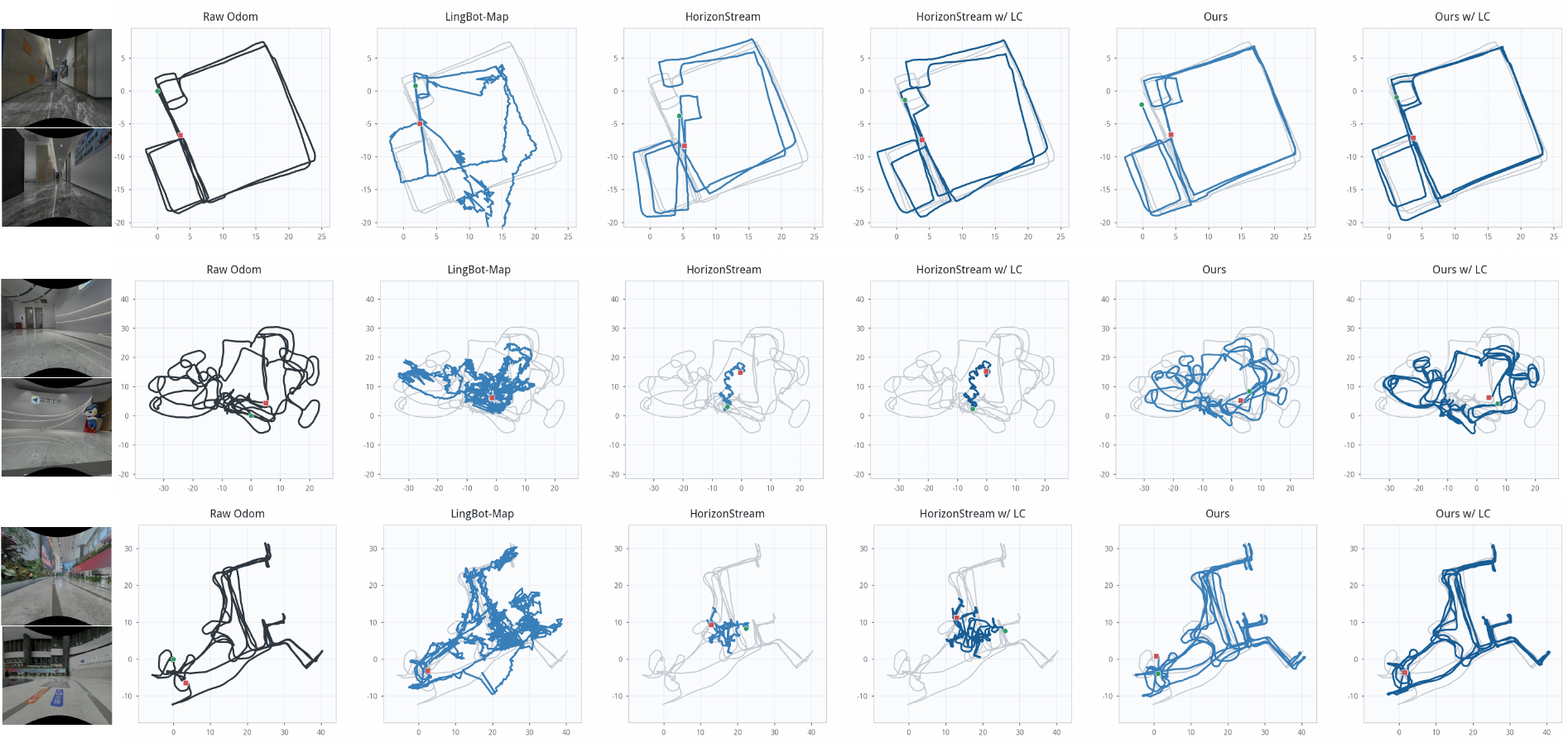}
  \caption{
  Qualitative camera trajectories on representative quadruped-robot
  sequences. The low camera height results in ground-dominant observations and motion patterns that differ from the training
  distribution.
  }
  \label{fig:vln_dog_results}
\end{figure*}

\paragraph{Low-viewpoint robot navigation}
Figure~\ref{fig:vln_dog_results} presents representative results on videos
captured by a quadruped robot. This setting differs noticeably from our
training data. The camera is mounted close to the ground, causing the ground
plane to occupy a large portion of the image while reducing the visibility of distant scene structures, making
frame-to-frame correspondence more challenging than in conventional handheld
or vehicle-mounted videos.

Despite this domain shift, our method produces more coherent trajectories across robot-navigation sequences than other methods. These examples provide qualitative evidence that our method can maintain
stable trajectory estimates under camera viewpoints and motion patterns that
are less frequently represented in the training data.

\begin{figure*}[t]
  \centering
  \includegraphics[width=\textwidth]{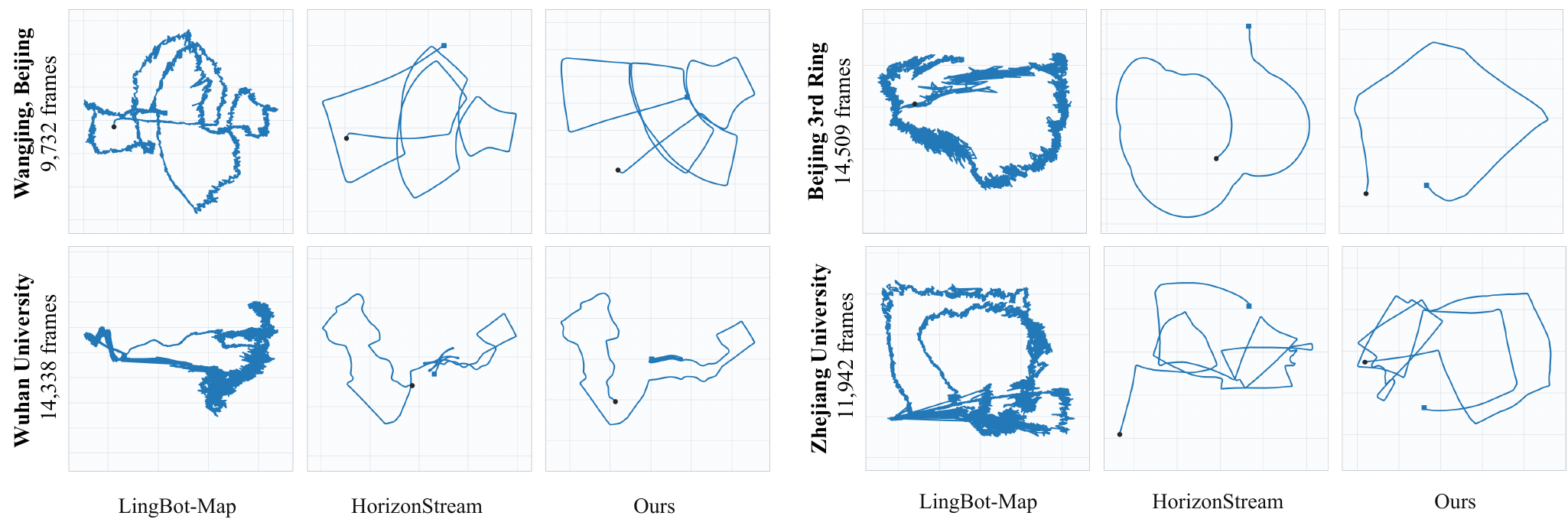}
\caption{
Qualitative no-loop trajectory comparisons on four large-scale real-world videos.
For efficiency and fair comparison, all methods are evaluated with temporal stride 5.
The results illustrate the behavior of streaming models under variations in scene scale, camera motion, and data distribution.
}
  \label{fig:large_scale_results}
\end{figure*}

\paragraph{Large-scale real-world sequences}
We additionally evaluate on four long videos as shown in
Figure~\ref{fig:large_scale_results}. These sequences cover different scene
scales and capture distributions, including extended urban driving, campus
roads, repeated structures, intersections, and substantial variations in
camera motion and visual appearance. They therefore complement the standard
benchmarks with less curated and more heterogeneous long-video streams.

Across the four sequences, our method produces trajectories with more
consistent large-scale structure than the compared streaming baselines. This
behavior is consistent with the motivation of our design: the network always
predicts an adjacent relative transform, so its regression target does not
change as the explored area or elapsed sequence length increases. The
lightweight temporal correction further suppresses locally correlated
rotational errors before they accumulate through pose composition.

Taken together, these qualitative results extend the benchmark evaluation to
camera viewpoints, motion patterns, and spatial scales. While accurate ground truth is unavailable for quantitative
assessment, the observed trajectory consistency suggests that keeping pose
prediction local provides a practical basis for generalization across
heterogeneous long-video streams.

\end{document}